\documentclass[authoryear]{kinamind}

\usepackage{multirow}
\usepackage{colortbl}
\usepackage{float}
\usepackage{placeins}

\definecolor{cgrn}{HTML}{D9F2D9}
\definecolor{cred}{HTML}{F8D7D7}
\definecolor{cyel}{HTML}{FFF3CC}
\definecolor{cgry}{HTML}{E8E8E8}

\title{Do AI Personas Grow? Analyzing and Benchmarking Personality Evolution in LLM Agents After Life Events}
\shorttitle{Do AI Personas Grow?}

\author[1,2]{Ming Wang\orcid{0000-0001-8406-5677}}
\author[1]{Peidong Wang\orcid{0009-0008-5015-1065}}
\author[1]{Xiaocui Yang\orcid{0000-0001-5352-8579}}
\author[1]{Daling Wang\orcid{0000-0003-1340-0778}}
\author[1]{Shi Feng\orcid{0000-0002-2846-7652}}
\author[2]{Fiona Fui-Hoon Nah\orcid{0000-0002-5505-7843}}
\author[2]{Ee-Peng Lim\orcid{0000-0003-0065-8665}}
\affil[1]{School of Computer Science and Engineering, Northeastern University, Shenyang, Liaoning, China}
\affil[2]{School of Computing and Information Systems, Singapore Management University, Singapore}

\correspondence{\href{mailto:wangdaling@cse.neu.edu.cn}{wangdaling@cse.neu.edu.cn};
\href{mailto:eplim@smu.edu.sg}{eplim@smu.edu.sg}}

\repo{github.com/sci-m-wang/BFI-Adapt}
\venue{KinaMind Society Preprint for arXiv}

\begin{document}
\maketitle

\begin{abstract}
Personality-conditioned LLM agents (PC-Agents) are increasingly used in emotional support, social simulation, and role-playing, motivating the development of lifelong agents that remain coherent over extended interactions. A key component of such coherence is personality evolution: agents should undergo plausible, psychology-grounded changes as they experience life events in different contexts. Although prior work shows that LLM personalities can shift under contextual perturbations, how these shifts vary across traits, events, personas, and models remains poorly understood. We study event-induced personality change after 11 major life events, using the Big Five traits as a psychometric anchor and interpreting the resulting trajectories against longitudinal evidence from human personality psychology. Across four diagnostic axes, PC-Agents exhibit measurable trait shifts at similar rates for event--trait pairs with and without documented human change directions. Even when shifts follow the expected direction, their magnitudes usually fall below human effect-size ranges. Gender and cultural-region prompts show little moderating effect, while persona-level dispersion is compressed three- to four-fold relative to human samples. To enable systematic comparison, we introduce \textsc{BFI-Adapt}, a reusable benchmark for scoring the directional fidelity of event-induced personality change, and use it to rank 14 models. Across the full persona--event grid, a validation suite shows that the measured shifts exceed no-event retest noise, remain stable under independently paraphrased prompts, exhibit limited and model-dependent convergence with scenario-based behavioral choices, and persist across intervening unrelated dialogue. Together, these checks establish the measured trajectories as robust event-conditioned response patterns. Our results suggest that current PC-Agents simulate the mean of human personality dynamics, but not its shape. Code and evaluation resources are available at \url{https://github.com/sci-m-wang/BFI-Adapt}.
\end{abstract}

\section{Introduction}
\label{sec:intro}

Personality-conditioned LLM agents (PC-Agents) have become a foundational primitive across a growing list of applications \citep{chen2026systematicanalysisimpactpersona}. They power emotional companionship and mental-health support chatbots \citep{11503093}, populate social simulations \citep{10.1145/3800683} for behavioral research \citep{park2024generative1000,larooij2025validation}, and serve as long-horizon role-playing engines for interactive fiction, games, and digital tutoring \citep{park2023generative,chen2024oscars}. Multi-session systems such as AnnaAgent already couple evolving emotional and cognitive states with persistent memory in psychological counselling \citep{wang-etal-2025-annaagent}. Across these settings, \textbf{lifelong agents} must maintain a coherent persona across extended interactions, long wall-clock durations, and open-ended user-driven narrative arcs.
A foundational question for lifelong PC-Agents is how their personality evolves relative to humans. In human psychology, personality traits, though enduring, can change due to life events. We study changes in Big Five traits in one's development and in response to consequential experiences, with major life events providing some of the best-studied anchors. Job entry and promotion are followed by increases in Conscientiousness \citep{roberts2006patterns,buhler2024life}. Chronic illness and unemployment are associated with increases in Neuroticism \citep{specht2011stability,boyce2015personality}. Retirement is followed by a documented Conscientiousness decline \citep{schwaba2019retirement}. A PC-Agent that ignores such dynamics may remain locally consistent yet become globally implausible over long-term interaction, like a static script wearing a person's name rather than a persona that develops through experience.

A growing concurrent line of work has begun examining whether LLM personalities remain stable or change under temporal and contextual perturbations. \citet{bodroza2024personality} report limited temporal stability and a prosocial-leaning profile across seven LLMs when personality inventories are re-administered. \citet{yu2026ptcbench} introduce PTCBench, which exposes LLM personalities to external conditions, including locations and life events, and measures aggregate trait shifts under the NEO Five-Factor Inventory. These studies show that LLM personalities can shift. Lifelong PC-Agents therefore require an evolving personality trajectory that remains coherent with the persona and plausible with respect to human psychological development. Several key properties remain unknown, including whether these shifts are systematic, whether they differ across traits and events, whether they preserve demographic and individual-level variation, and whether their direction and magnitude resemble patterns documented in human personality-change evidence. These unknowns translate into three design requirements for our diagnostic study. First, because systematic and idiosyncratic shifts can cancel in aggregate means, we move to event--trait-level diagnostics, including item-level reliability and directional consistency. Second, because direction and magnitude must be judged against an external psychological reference, we anchor each event--trait pair on an expected direction of trait change for that event, drawn from meta-analytic and primary scientific evidence \citep{buhler2024life,bleidorn2018life,roberts2006patterns,boyce2015personality,schwaba2019retirement}. Third, because persona heterogeneity is part of the target phenomenon, we instantiate each LLM with 100 demographically controlled personas, making it possible to test whether model responses preserve demographic moderation and individual-level dispersion observed in human samples.

We study 11 major life events drawn from the human longitudinal literature, covering occupational events, social events, and health events. For each event, we instantiate 100 demographically controlled personas in a factorial design of 2 genders $\times$ 5 cultural regions $\times$ 10 personality archetypes. Each persona first completes the BFI-44 inventory, then undergoes a structured event-reflection stage, and finally completes the same inventory again. We run this pipeline across 11 LLMs. To establish the robustness of the resulting trajectory signal, we additionally conduct no-event retests, independent event paraphrases, scenario-based decision measurements, and delayed remeasurement after unrelated dialogue. The analysis is structured around four research questions, each examining a different axis of PC-Agent personality evolution against expected human change patterns.

\begin{itemize}[noitemsep, topsep=0pt, leftmargin=*]
    \item \textbf{RQ1 (Existence).} Do PC-Agents show personality change after life events, in the sense that personas leave the BFI subscale noise floor?
    \item \textbf{RQ2 (Direction and magnitude).} When PC-Agents move, do they match the expected direction and magnitude of human personality change for that event--trait pair?
    \item \textbf{RQ3 (Demographic shape).} Do trait change patterns vary across persona gender and cultural strata in a way that resembles the moderator structure reported in human studies?
    \item \textbf{RQ4 (Individual shape).} Within a single event--trait pair, do PC-Agents produce persona-specific variation comparable to the population-level dispersion in human samples, or do they collapse personas onto a shared mean trajectory?
\end{itemize}

\begin{figure*}[t]
	\centering
	\includegraphics[width=\textwidth]{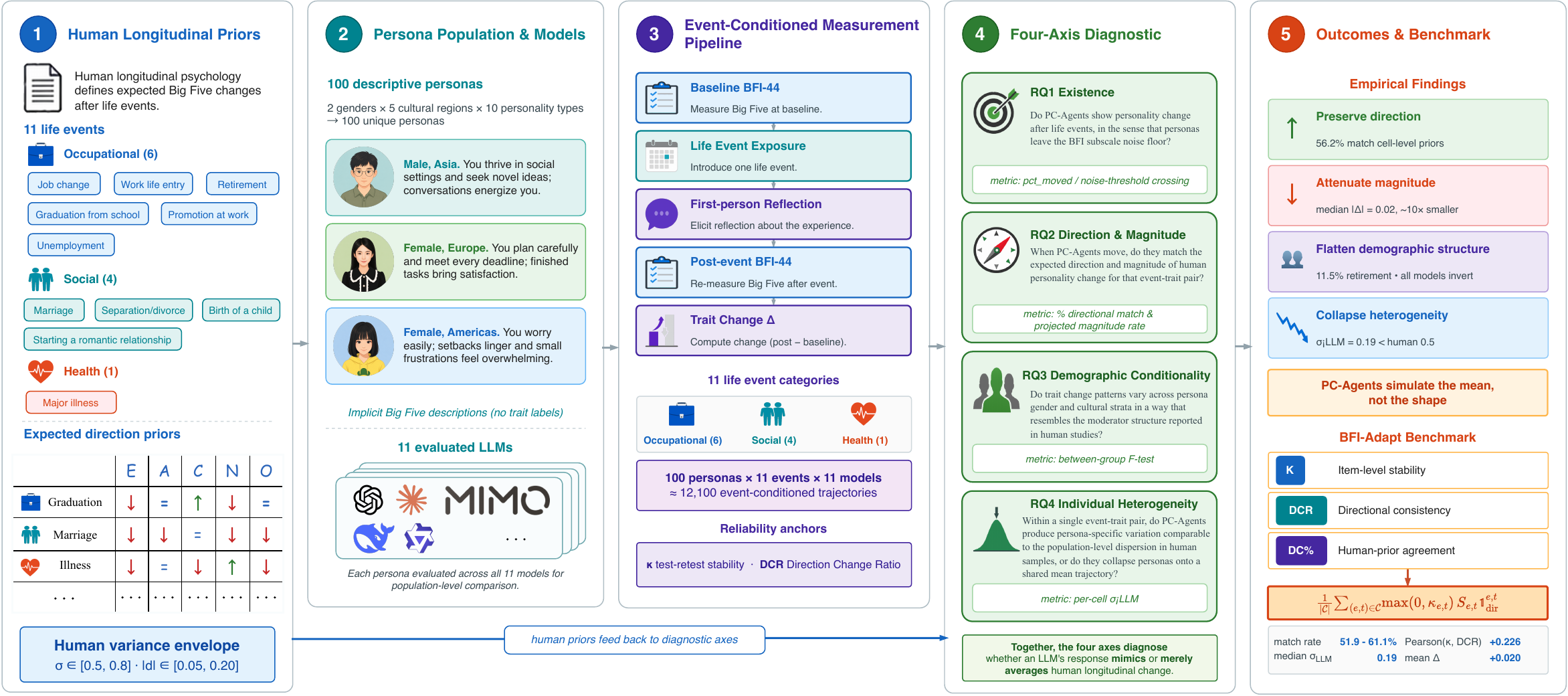}
	\caption{Overview of our analytical framework for studying personality evolution in PC-Agents.}
	% We expose demographically controlled personas to life events, measure pre- and post-event BFI-44 profiles, and compare the resulting trait changes against human change-direction priors along four axes: existence (whether ratings cross the BFI noise floor), direction and magnitude (signed projected $\Delta$ vs.\ human envelope), demographic shape (variability across gender and culture strata), and individual shape (persona-level dispersion within a pair). Event-level direction priors are adopted from \citet{SPECHT2017341} (Table~\ref{tab:expected}); magnitude and dispersion envelopes are calibrated from independent meta-analytic evidence \citep{buhler2024life,bleidorn2018life}. The same measurements are later packaged into \textsc{BFI-Adapt}, a reusable composite over item-level reliability, directional consistency, and prior-aligned direction.
	\label{fig:framework}
	% \vspace{-0.5em}
\end{figure*}

Across the four research questions, the central observation is consistent. PC-Agents move, but their movement is weakly event-specific, poorly calibrated in magnitude, and compressed across demographic and persona-level variation. They therefore simulate the mean of human personality dynamics more readily than its shape. The validation suite shows that event-conditioned changes exceed retest noise, preserve their event--trait structure under independent paraphrases, exhibit model-dependent convergence with scenario-based decisions, and remain detectable after unrelated dialogue. Building on the RQ2 diagnostic, we introduce \textsc{BFI-Adapt}, an evaluation method for scoring whether event-induced personality shifts follow expected human directions. Figure~\ref{fig:framework} provides an overview. We make three contributions.

\begin{itemize}[noitemsep, topsep=0pt, leftmargin=*]
	\item We frame PC-Agent personality evolution as a human-prior diagnostic over four axes: existence, direction and magnitude, demographic shape, and individual shape.
    \item We benchmark diverse LLMs with controlled personas and paired BFI-44 measurements, revealing indiscriminate movement, magnitude miscalibration, demographic invariance, and heterogeneity collapse. We validate the resulting trajectory analysis through no-event retests, independent paraphrases, scenario-based decisions, and short-range retention.
    \item We introduce \textsc{BFI-Adapt}, an evaluation method for measuring whether LLM-agent personality shifts adapt to expected directions exhibited by humans, and rank 14 models with it.
\end{itemize}

\section{Related Work}
\label{sec:related}

% \paragraph{Personality Change After Life Events.}
% A robust body of longitudinal research documents systematic Big Five personality changes following major life events. \citet{roberts2006patterns} established through meta-analysis that personality traits show normative mean-level changes across the lifespan. \citet{specht2011stability} demonstrated that specific life events---including changes in employment status, relationship transitions, and health crises---produce measurable shifts in Big Five traits. \citet{bleidorn2018life} reviewed the bidirectional relationship between life events and personality change, finding small but reliable effects (Cohen's $d < 0.5$). The recent meta-analysis of \citet{buhler2024life} pools 89 samples ($N = 121{,}187$) and confirms that graduation, the first job, new relationships, marriage, and divorce produce the strongest and most consistent trait changes, while \citet{boyce2015personality} document that unemployment lowers agreeableness, conscientiousness, and openness, and \citet{schwaba2019retirement} report a Conscientiousness decline following retirement.

% \subsection{Static LLM Personality Assessment.}
\paragraph{Static LLM Personality Assessment.}
% A first line of work examines whether LLMs can express stable personality profiles under standard inventories or persona prompts. 
Prior studies show that LLMs can produce reliable Big Five profiles, differentiate prompted personality levels, and support linguistic or inventory-based personality assessment \citep{safdari2023personality,jiang2024personallm,zheng2025lmlpa,handa2025personality}. These studies establish the feasibility of measuring static LLM personality, but largely treat personality as a fixed trait profile rather than a trajectory that changes after life events.
Extending beyond a single inventory, \citet{huang2024humanity} evaluate five LLMs with thirteen clinical-psychology scales spanning personality, relationships, motivation, and emotional ability. Moving beyond self-report inventories, \citet{wang-etal-2026-genpt} adapt projective tests to LLMs and find improved resistance to contamination and social-desirability bias, together with greater sensitivity to prompt-induced changes. They further demonstrate GenPT in longitudinal counselling. These results motivate our use of BFI-44 as a standardized and interpretable measurement anchor for systematic, event-conditioned trajectories.

% \subsection{Role-Playing Agents.}
\paragraph{Role-Playing Agents.}
Role-playing agents provide another route to personality-conditioned behavior, with benchmarks and training methods evaluating whether models remain faithful to a specified character or persona \citep{wang2024rolellm,wang2024incharacter,ran2024capturing}. Recent work also shows that user personas can shift chatbot personality \citep{xing2025chameleon}, while induced personas yield stable, task-dependent changes in cognitive performance \citep{chen2026systematicanalysisimpactpersona}. However, this line of work primarily targets persona fidelity, whether the model stays in character, rather than whether the in-character personality evolves in psychologically plausible ways.
CharacterEval, for example, evaluates role-playing conversations with thirteen metrics over four dimensions and a multi-turn benchmark of 77 characters \citep{tu2024charactereval}. Such work clarifies whether an agent preserves its assigned character, whereas our controlled personas provide fixed initial conditions from which change is measured. We retain persona conditioning throughout and ask whether subsequent adaptation is event-specific and comparable to human change patterns.

\paragraph{Dynamic LLM Personality.}
Recent studies have begun testing whether LLM personalities remain stable or shift under temporal and contextual perturbations. \citet{bodroza2024personality} re-administer personality inventories to seven LLMs and report limited temporal stability and a prosocial-leaning profile. \citet{yu2026ptcbench} introduce PTCBench, exposing LLMs to external conditions, including locations and life events, and measuring aggregate NEO-FFI trait shifts. Complementing these inventory-based findings, \citet{han2025personality} compare LLM self-reports with behavioral tasks and find that persona interventions steer reported traits more reliably than behavior. Together, these works show that LLM personality can change, but they provide limited analysis of how such changes unfold across traits, events, personas, and models, or whether the measured shifts remain consistent across complementary checks. Our work addresses these gaps by moving from aggregate trait shifts to item-level reliability, anchoring event--trait pairs in human change-direction priors, and testing demographic conditionality and individual-level heterogeneity with 100 personas per model. We further evaluate trajectory robustness through retest comparisons, independent paraphrases, scenario-based decision behavior, and short-range retention.

% \paragraph{Generative Social Simulation.}
% \citet{park2023generative} demonstrated that LLM-powered generative agents can produce believable social behaviors. \citet{park2024generative1000} scaled this to simulate 1{,}000 individuals with measured personality accuracy. \citet{larooij2025validation} argued that validation against human behavioral data is the central challenge for generative social simulations. Our work directly addresses this challenge for the specific dimension of event-driven personality dynamics.

\section{Methodology}
\label{sec:method}

\subsection{Persona Design}
\label{sec:personas}

To isolate personality dynamics from confounding factors inherent in fictional characters, such as narrative expectations, cultural stereotypes, and fan interpretations, we created 100 descriptive personas using a factorial design. It crosses 2 genders (male, female), 5 cultural regions (Europe, Americas, Africa, Asia, Oceania), and 10 personality types (P1--P10). The 10 personality types are constructed from implicit Big Five design anchors. We use a five-level ordinal scheme with ``very low,'' ``low,'' ``moderate,'' ``high,'' and ``very high'' levels for persona construction. Each personality type assigns one focal Big Five trait to either ``very high'' or ``very low,'' while keeping the remaining four traits ``moderate.'' Thus, the 10 types correspond to the very-high and very-low variants of five traits. These ordinal anchors guide the writing of behavioral and attitudinal persona descriptions, such as ``thrives in social gatherings and draws energy from conversation.'' The BFI-44 administrations provide the standard 1--5 Big Five scores used in all analyses.

\subsection{Life Events}
\label{sec:events}

We selected 11 life events associated with Big Five personality changes in the psychology literature. They span occupational events (graduation, work entry, job change, promotion, unemployment, retirement), social events (new relationship, marriage, divorce, childbirth), and health events (chronic illness). To define the expected direction of change for each event and trait, we use \citet{SPECHT2017341} as the primary source and consolidate its reported directions into the event--trait prior matrix in Table~\ref{tab:expected}. We make one notational modification. Emotional Stability (ES) is inverted to Neuroticism (N) to match our BFI-44 scoring convention ($\text{ES}{+}\!\equiv\!\text{N}{-}$). Independent meta-analytic and primary studies support the individual rows for job entry and promotion \citep{roberts2006patterns,buhler2024life}, unemployment \citep{boyce2015personality}, retirement \citep{schwaba2019retirement}, and chronic illness \citep{specht2011stability}. %Table~\ref{tab:expected} reproduces the resulting expected directions.

\begin{table}[!b]
	\centering
	\small
	\caption{Expected directions of human Big Five trait change across 11 life events. ``--'' denotes no strong change. ``$\pm$'' denotes a direction that depends on role demand.}
	\label{tab:expected}
	\begin{tabular}{lccccc}
		\toprule
		\textbf{Event} & \textbf{E}   & \textbf{A}   & \textbf{C}   & \textbf{N}   & \textbf{O}   \\
		\midrule
		Graduation     & $\downarrow$ & --           & $\uparrow$   & $\downarrow$ & --           \\
		Work Entry     & --           & --           & $\uparrow$   & $\downarrow$ & --           \\
		Job Change     & $\pm$        & --           & --           & --           & $\pm$        \\
		Promotion      & --           & --           & $\uparrow$   & $\downarrow$ & $\uparrow$   \\
		Unemployment   & --           & $\downarrow$ & $\downarrow$ & --           & $\downarrow$ \\
		Retirement     & --           & --           & $\downarrow$ & --           & --           \\
		New Rel.       & $\uparrow$   & $\downarrow$ & --           & $\downarrow$ & --           \\
		Marriage       & $\downarrow$ & $\downarrow$ & --           & $\downarrow$ & $\downarrow$ \\
		Divorce        & --           & $\uparrow$   & --           & --           & $\uparrow$   \\
		Child Birth    & $\downarrow$ & --           & $\downarrow$ & --           & --           \\
		Chronic Ill.   & $\downarrow$ & --           & $\downarrow$ & $\uparrow$   & $\downarrow$ \\
		\bottomrule
	\end{tabular}
\end{table}

\subsection{Personality Assessment Pipeline}
\label{sec:pipeline}

We design a four-stage pipeline to measure personality change in a model that simulates a persona experiencing a life event. The system message specifies gender, cultural region, and a behavioral personality description. The model answers all subsequent prompts in character. \textbf{Baseline Measurement} administers the BFI-44 inventory \citep{john2008paradigm} and yields a Big Five score vector $\mathbf{b}$, where $\mathbf{b}_t \in [1,5]$ and $t \in \{E, A, C, N, O\}$. \textbf{Event Presentation} gives the simulated persona a life event and elicits a first-person reflection. \textbf{Post-Event Measurement} administers BFI-44 again and yields the score vector $\mathbf{p}$. \textbf{Change Computation} calculates $\Delta_t = \mathbf{p}_t - \mathbf{b}_t$ for each trait. The 44 items use a 1--5 Likert scale with standard reverse scoring. Trait scores average the corresponding items, with 8 items for E, 9 for A, 9 for C, 8 for N, and 10 for O. The resulting trait-specific step sizes are 0.125 for E and N, 0.111 for A and C, and 0.1 for O. %Each persona is measured once at baseline and once after each of the 11 events. The per-persona baseline is reused across the 11 event conditions because it is collected before any event exposure and therefore provides a common pre-event anchor for that persona.

\subsection{Statistical Analysis}
\label{sec:stats}

We evaluate change direction at the persona level because aggregate means can cancel opposing individual shifts. A mean near zero can coexist with directional effects across personas. We therefore adopt an individual-level direction classification approach.

\paragraph{Direction Classification.}
For each persona $p$, event $e$, and trait $t$, we classify the change direction:
\begin{equation}
	d_{p,e,t} = \begin{cases}
		+ & \text{if } \Delta_{p,e,t} > \varepsilon      \\
		- & \text{if } \Delta_{p,e,t} < -\varepsilon     \\
		0 & \text{if } |\Delta_{p,e,t}| \leq \varepsilon
	\end{cases}
	\label{eq:direction}
\end{equation}
where $\varepsilon = 0.1$ serves as a conservative noise boundary. A shift exactly at $\pm0.1$ remains neutral. Shifts beyond that boundary count as directional, and neutral cases are excluded from the analysis denominator.

\paragraph{Match Rate.}
Suppose $P$ denotes a set of personas. For each event--trait pair $(e,t)$ with an expected human direction $d_{e,t}^{\mathrm{human}}$, the match rate is defined by:
\begin{equation}
	P_{\text{match}}(e,t) = \frac{|\{p \in P : d_{p,e,t} = d_{e,t}^{\mathrm{human}}\}|}{|\{p \in P : d_{p,e,t} \neq 0\}|}
\end{equation}
$P_{\text{match}}(e,t)$ is therefore the conditional probability that a persona's non-neutral shift points in the expected human direction for that event--trait pair. The metric is defined for pairs with a definite expected human direction, with neutral responses excluded from the denominator.

\paragraph{Statistical Testing.}
The statistical tests evaluate whether the observed directional evidence exceeds a minimal chance baseline and whether it varies across demographic strata. For each definite-prior event--trait pair, the directional null is $H_0: P_{\text{match}}(e,t)=0.5$. We apply a one-tailed binomial sign test with Wilson-score confidence intervals and Benjamini--Hochberg FDR correction at $\alpha{=}0.05$. Cross-group comparisons for gender and world-region effects use Fisher's exact tests on $2\times2$ match/mismatch tables, with Benjamini--Hochberg correction within each test family. Appendix~\ref{app:stats} presents the full procedure.

\subsection{Item-Level Reliability: $\kappa$ and DCR}
\label{sec:kappa_dcr}

Match rate captures the alignment of a moved persona with the expected direction. Two additional indicators characterize the underlying item responses at the event--trait-pair level. Linearly weighted Cohen's $\kappa$ measures item-level rating stability between baseline and post-event answers. DCR measures whether the residual item changes within the same pair are mostly one-sided. Pair-level computation preserves the distinct human direction associated with each event--trait pair and prevents opposing directions from cancelling during aggregation.

\paragraph{Pair-level linearly-weighted Cohen's $\kappa$.} For each model and event--trait pair $(e,t)$, let $n_t$ be the number of BFI-44 items belonging to trait $t$ (Section~\ref{sec:setup}). The pair contains $100\,n_t$ paired item ratings (100 personas $\times$ $n_t$ subscale items, $K{=}5$ ordinal categories). With weights $w_{ij} = 1 - |i{-}j|/(K{-}1)$,
\begin{equation}
	\kappa_{e,t} = \frac{\sum_{i,j} w_{ij}\, p^{(o)}_{ij} \;-\; \sum_{i,j} w_{ij}\, p^{(b)}_i p^{(p)}_j}{1 \;-\; \sum_{i,j} w_{ij}\, p^{(b)}_i p^{(p)}_j} \in [-1, 1],
	\label{eq:kappa}
\end{equation}
where $p^{(o)}_{ij}$ is the empirical joint frequency of baseline rating $i$ and post-event rating $j$ within the pair and $p^{(b)}_i, p^{(p)}_j$ are the corresponding marginals. $\kappa_{e,t}{=}1$ is perfect agreement, $\kappa_{e,t}{=}0$ is chance-level agreement. We summarize each model by $\bar{\kappa}_{\text{pair}}$, the mean of $\kappa_{e,t}$ over the 27 event--trait pairs with a definite expected human direction.

\paragraph{Pair-level Directional Consistency Ratio (DCR).} Within the same event--trait pair, let $n_\uparrow$ and $n_\downarrow$ be the number of items whose post-event rating increases and decreases (reverse-coded items pre-aligned), respectively. Among the $n_\uparrow + n_\downarrow$ items that change,
\begin{equation}
	\mathrm{DCR}_{e,t} = \frac{\max(n_\uparrow, n_\downarrow)}{n_\uparrow + n_\downarrow} \in [0.5, 1].
	\label{eq:dcr}
\end{equation}
If no item changes within a pair, $\mathrm{DCR}_{e,t}$ has no empirical denominator. For model-level summaries we report such pairs as $\mathrm{DCR}_{e,t}{=}0.5$, equivalent to no dominant direction and hence $S_{e,t}{=}2\,\mathrm{DCR}_{e,t}{-}1{=}0$ in \textsc{BFI-Adapt}. We summarise each model by $\overline{\mathrm{DCR}}_{\text{pair}}$, the mean of $\mathrm{DCR}_{e,t}$ over the same 27 definite-direction pairs. Thus, $\kappa_{e,t}$ measures the similarity of item ratings before and after the event, while $\mathrm{DCR}_{e,t}$ measures whether changed ratings move primarily in one direction. Pooled forms $\kappa_{\text{pool}}$ and $\mathrm{DCR}_{\text{pool}}$ are reported in Table~\ref{tab:kappa_dcr_legacy} to show the artefact created by aggregating across pairs with opposite expected directions. Appendix~\ref{app:reliability} details the four diagnostic regimes formed by the $(\kappa_{e,t}, \mathrm{DCR}_{e,t})$ plane.

\subsection{Validation Design}
\label{sec:validation_design}

We evaluate the robustness of the event-conditioned trajectory signal through four complementary designs. A no-event BFI-44 retest establishes a model-specific measurement floor. Independent event paraphrases test whether the event--trait structure remains stable under new wording. Counterbalanced parallel forms of ten scenario-based decisions measure convergence between BFI trait changes and concrete choices. A delayed BFI-44 administration after three unrelated dialogue turns measures short-range retention. Each condition starts from a fresh conversation. Confidence intervals use persona-cluster bootstrap resampling. Additional prompts, counterbalancing procedures, and statistical details appear in Appendix~\ref{app:validity}.

\section{Experiments}

\subsection{Experiment Settings}
\label{sec:setup}

\paragraph{Models.}
The primary four-axis analysis covers 11 LLMs. These are GPT-5.3-chat \citep{singh2026openaigpt5card}, GPT-4.1-mini \citep{openai2024gpt4technicalreport}, Claude-Sonnet-4.6, Claude-Haiku-4.5, Gemini-3-flash \citep{geminiteam2025geminifamilyhighlycapable}, DeepSeek-V4-Pro \citep{deepseekai2026deepseekv4}, Doubao-Seed-2.0-Pro, MiMo-V2.5-Pro \citep{mimo2026v25pro}, Qwen3-235B \citep{yang2025qwen3technicalreport}, GLM-4.6 \citep{5team2025glm45agenticreasoningcoding}, and Kimi-K2 \citep{kimiteam2026kimik2openagentic}. Three open-weight models, Qwen3.5-9B, Qwen2.5-14B-Instruct, and InternLM3-8B-Instruct, run the same grid and extend the \textsc{BFI-Adapt} leaderboard to 14 models. The validation suite evaluates DeepSeek-V4-Pro, GPT-5.5, Grok-4.3, Kimi-K2.6, Mistral-Large-3, Qwen3.5-9B, Qwen2.5-14B-Instruct, and InternLM3-8B-Instruct. All models use non-reasoning mode. Each main-grid model completes the full $100$ personas $\times$ $11$ events $\times$ $44$ items design. The validation models use the same 100 personas and 11 events under each validation condition. Appendix~\ref{app:reliability} reports response-integrity checks.

\paragraph{Prompt.}
The system prompt establishes gender, continent, and a behavioural personality description that never names Big Five labels. Appendix~\ref{app:prompts} shows the exact persona-simulation prompt, and Appendix~\ref{app:persona} gives one inserted personality description. BFI-44 items are answered in character. Event scenarios appear as first-person narratives that the persona reflects on before the post-event items. The prompts are available at \url{https://github.com/sci-m-wang/BFI-Adapt}.

\subsection{RQ1: Existence}
\label{sec:rq1}

RQ1 asks whether PC-Agents produce measurable personality responses after life events, before asking whether such responses are psychologically targeted. We define movement at the persona level as $|\Delta|>0.1$, a conservative threshold just above the smallest BFI-44 trait-level step, and compute $ \mathrm{pct}_{\text{moved}}(m,e,t)=|\{p:|\Delta_{p,e,t}|>0.1\}|/100 $ for each model--event--trait pair.

We compute this statistic over all 605 model--event--trait combinations and, within each model, compare 27 pairs with documented human change directions against 28 pairs without a definite direction. Within-model median differences quantify the separation between these two groups. Appendix Table~\ref{tab:existence_per_model} reports per-model quartiles and high-tail mass $\mathrm{Hi}\%$ ($\mathrm{pct}_{\text{moved}}>0.6$). Movement occurs broadly in both groups. Definite-direction medians span 0.44--0.84, no-definite-direction medians span 0.42--0.82, and within-model median differences remain below 0.05. In 9 of 11 models, the two $\mathrm{Hi}\%$ values differ by less than 10 percentage points. The two larger differences favor no-definite-direction pairs. PC-Agent movement is therefore widespread but weakly targeted toward pairs with human directional evidence.
As a reliability check for the subsequent direction analysis, we also inspect pair-level $\bar{\kappa}_{\text{pair}}$ on the 27 definite-direction pairs. It spans 0.58--0.86 across the 11-model benchmark (Table~\ref{tab:kappa_dcr_per_model}). 10 of the 11 models exceed $\bar{\kappa}_{\text{pair}}{=}0.66$. MiMo-V2.5-Pro is the single low-reliability outlier at 0.58 and also the model with near-universal movement ($\mathrm{Hi}\%{=}1.00$ in both buckets). High $\bar{\kappa}_{\text{pair}}$ demonstrates reliable item-level ratings within each pair. More than half of personas still cross the noise floor on a typical pair. Together, these properties support the direction and magnitude analysis in Section~\ref{sec:rq2_dir_mag}.

% Auto-generated by scripts/kappa_dcr_analysis.py — DO NOT EDIT
\begin{table}[t]
\centering\small
\caption{Pair-level reliability, directional consistency, and the \textsc{BFI-Adapt} composite for the 14-model PC-Agent leaderboard. Scores are computed over the 27 event--trait pairs with a definite expected human direction. The last three rows form the open-weight extension.}
\label{tab:kappa_dcr_per_model}
\begin{tabular*}{\columnwidth}{l@{\extracolsep{\fill}}rrrr}
\toprule
Model & $\bar{\kappa}_{\text{pair}}$ & $\overline{\mathrm{DCR}}_{\text{pair}}$ & $\mathrm{DC}\%_{\text{pair}}$ & \textsc{Adapt} \\
\midrule
Gemini-3-flash & 0.862 & 0.811 & 59.3 & 0.348 \\
GLM-4.6 & 0.825 & 0.804 & 63.0 & 0.322 \\
Qwen3-235B & 0.734 & 0.791 & 63.0 & 0.287 \\
Claude-Haiku-4.5 & 0.829 & 0.697 & 66.7 & 0.225 \\
Kimi-K2-0905 & 0.837 & 0.702 & 70.4 & 0.224 \\
Doubao-Seed-2.0-Pro & 0.767 & 0.771 & 48.1 & 0.218 \\
Claude-Sonnet-4.6 & 0.860 & 0.707 & 55.6 & 0.212 \\
GPT-4.1-mini & 0.842 & 0.666 & 55.6 & 0.192 \\
GPT-5.3-chat & 0.814 & 0.704 & 51.9 & 0.178 \\
DeepSeek-V4-Pro & 0.793 & 0.691 & 55.6 & 0.176 \\
MiMo-V2.5-Pro & 0.576 & 0.598 & 63.0 & 0.071 \\
\midrule
Qwen3.5-9B & 0.793 & 0.739 & 63.0 & 0.194 \\
Qwen2.5-14B-Instruct & 0.522 & 0.571 & 55.6 & 0.049 \\
InternLM3-8B-Instruct & 0.515 & 0.563 & 63.0 & 0.044 \\
\bottomrule
\end{tabular*}
\vspace{-0.5em}
\end{table}
 
\subsection{RQ2: Direction and Magnitude}
\label{sec:rq2_dir_mag}

RQ2 asks two linked questions. When a persona moves, does it move in the expected human direction? Is the signed magnitude comparable to human longitudinal effects? We first use pair-level DCR to characterize directional item changes within each event--trait pair. We then use $P_{\text{match}}$ and $\mathrm{DC}\%_{\text{pair}}$ to measure agreement with the expected direction from human longitudinal studies. Per-pair $P_{\text{match}}$ is tested against the $50\%$ chance baseline via Wilson 95\% CIs with BH-FDR at $q{=}0.05$ (Appendix~\ref{app:stats}). $\overline{\mathrm{DCR}}_{\text{pair}}$, averaged over the 27 definite-direction pairs, spans 0.60--0.81 across the 11 models, with every model above 0.59. Residual item drift is therefore usually one-sided within a pair. Directional agreement is weaker. $\mathrm{DC}\%_{\text{pair}}$ spans 48.1--70.4\%, from Doubao-Seed-2.0-Pro at the bottom to Kimi-K2-0905 at the top. Stability and systematicity are modestly correlated across models with $r{=}{+}0.455$, confirming that they capture complementary properties.

At finer resolution, the model-by-event heatmap (Figure~\ref{fig:match_heatmap}, Appendix~\ref{app:supp_figures}) shows three recurring patterns. First, occupational onboarding events are easiest: \textsc{Graduation} (42--76\%), \textsc{Work Entry} (49--96\%), and \textsc{Promotion} (53--86\%) are correctly directed by most models, consistent with the well-represented occupational-role $\to$ Conscientiousness association \citep{roberts2006patterns}. Second, \textsc{Retirement} breaks every model: the documented post-retirement Conscientiousness decline \citep{schwaba2019retirement} is universally reversed (0.0--38.9\%, median 11.5\%), with models predicting retirees to become more conscientious. Third, social events remain near chance (\textsc{Marriage} 44--74\%, \textsc{Divorce} 40--76\%, and \textsc{Child Birth} 25--63\%), whereas \textsc{Chronic Illness} is more consistently captured, likely because its expected profile is salient (E/C/O decline, N increases). Aggregating by trait, \textbf{Agreeableness is the weakest dimension} (30--58\% across models), below Openness (53--62\%), Conscientiousness (45--62\%), Extraversion (41--74\%), and Neuroticism (49--79\%), reflecting a default tendency to make personas more agreeable after major events even when human evidence expects the opposite.

For magnitude, we project each persona-level change onto the expected human direction, $\widetilde{\Delta}=\mathrm{sign}(d_{e,t}^{\mathrm{human}})\cdot\Delta$. The meta-analysis reports standardized mean change $d$ \citep{buhler2024life}, for which we use the representative band $|d|\in[0.05,0.20]$. Applying a representative BFI trait SD of $\sigma{\approx}0.7$ gives the raw-change reference band $\widetilde{\Delta}\in[0.035,0.14]$ Likert units. We classify responses as reversed ($\widetilde{\Delta}<0$), under-shift ($0\leq\widetilde{\Delta}<0.035$), in-range ($0.035\leq\widetilde{\Delta}\leq0.14$), or overshoot ($\widetilde{\Delta}>0.14$). Binning is performed separately for each model over its 2700 persona-level responses (100 personas $\times$ 27 definite-direction pairs). Across models, only \textbf{11.0\%--16.4\%} of responses fall inside the human reference band. The rest are reversed (\textbf{20.8\%--40.2\%}), under-shifted despite the correct direction (\textbf{15.1\%--54.0\%}), or overshooting (\textbf{9.9\%--31.6\%}). Failure modes differ by model. Kimi-K2-0905 most often under-shifts, MiMo-V2.5-Pro most often reverses or overshoots, and the best-calibrated model, Gemini-3-flash, reaches 16.4\% in-range. Thus, PC-Agents partly recover direction but rarely calibrate the magnitude of change to human effect-size ranges.

Table~\ref{tab:cell_grid} expands the comparison to all five traits per event. Each entry reports $\mathrm{pct}_{+}\,/\,\mathrm{pct}_{-}$, the cross-model median fraction of personas with $\Delta{>}0$ and $\Delta{<}0$. Definite-direction pairs are marked as match (green) or reverse (red) using a 10\% net directional-intensity threshold. No-definite-direction pairs are marked as drift (yellow) when over half the personas leave the noise floor. Among the 27 definite-direction pairs, 14 (51.9\%) match the expected direction and 13 (48.1\%) reverse it, with failures concentrated on retirement, unemployment, divorce, and new relationships. Among the 26 no-definite-direction pairs, 21 (80.8\%) drift, showing broad spillover onto bystander traits, around events that produce directional failures.

% auto-generated by scripts/rq_stats.py
\begin{table}[t]
  \centering\scriptsize
  \setlength{\tabcolsep}{1.6pt}
  \renewcommand{\arraystretch}{1.10}
  \caption{Event--trait direction patterns. Each cell shows the expected direction followed by $\mathrm{pct}_{+}\,/\,\mathrm{pct}_{-}$, the percentages of personas with post-event increases and decreases. Green marks a matched expected direction, red a reversal, yellow drift without a definite prior, white stability without a definite prior, and gray a context-dependent direction.}
  \label{tab:cell_grid}
  \begin{tabular}{l ccccc}
    \toprule
    \textbf{Event} & \textbf{E} & \textbf{A} & \textbf{C} & \textbf{N} & \textbf{O} \\
    \midrule
    Graduation & \cellcolor{cgrn} $\downarrow$\,17/34 &  --\,30/21 & \cellcolor{cgrn} $\uparrow$\,35/17 & \cellcolor{cgrn} $\downarrow$\,22/33 & \cellcolor{cyel} --\,28/39 \\
    Work Entry & \cellcolor{cyel} --\,17/36 & \cellcolor{cyel} --\,33/17 & \cellcolor{cgrn} $\uparrow$\,58/7 & \cellcolor{cgrn} $\downarrow$\,17/37 & \cellcolor{cyel} --\,22/46 \\
    Job Change & \cellcolor{cgry} $\pm$\,19/39 &  --\,27/21 & \cellcolor{cyel} --\,52/12 & \cellcolor{cyel} --\,23/38 & \cellcolor{cgry} $\pm$\,34/36 \\
    Promotion & \cellcolor{cyel} --\,27/29 & \cellcolor{cyel} --\,35/17 & \cellcolor{cgrn} $\uparrow$\,69/5 & \cellcolor{cgrn} $\downarrow$\,13/46 & \cellcolor{cred} $\uparrow$\,24/45 \\
    Unemployment & \cellcolor{cyel} --\,22/32 & \cellcolor{cred} $\downarrow$\,26/15 & \cellcolor{cred} $\downarrow$\,45/10 & \cellcolor{cyel} --\,43/17 & \cellcolor{cgrn} $\downarrow$\,20/48 \\
    Retirement & \cellcolor{cyel} --\,14/42 & \cellcolor{cyel} --\,38/16 & \cellcolor{cred} $\downarrow$\,64/7 & \cellcolor{cyel} --\,9/49 & \cellcolor{cyel} --\,13/60 \\
    New Rel. & \cellcolor{cred} $\uparrow$\,25/28 & \cellcolor{cred} $\downarrow$\,30/15 &  --\,25/20 & \cellcolor{cgrn} $\downarrow$\,13/33 & \cellcolor{cyel} --\,31/40 \\
    Marriage & \cellcolor{cred} $\downarrow$\,26/26 & \cellcolor{cred} $\downarrow$\,36/17 &  --\,35/15 & \cellcolor{cgrn} $\downarrow$\,9/42 & \cellcolor{cgrn} $\downarrow$\,19/50 \\
    Divorce & \cellcolor{cyel} --\,18/41 & \cellcolor{cred} $\uparrow$\,22/32 & \cellcolor{cyel} --\,34/20 & \cellcolor{cyel} --\,66/14 & \cellcolor{cred} $\uparrow$\,17/49 \\
    Child Birth & \cellcolor{cred} $\downarrow$\,23/33 &  --\,37/14 & \cellcolor{cred} $\downarrow$\,33/22 & \cellcolor{cyel} --\,37/21 & \cellcolor{cyel} --\,21/49 \\
    Chronic Ill. & \cellcolor{cgrn} $\downarrow$\,16/33 & \cellcolor{cyel} --\,25/25 & \cellcolor{cred} $\downarrow$\,33/17 & \cellcolor{cgrn} $\uparrow$\,49/14 & \cellcolor{cgrn} $\downarrow$\,16/53 \\
    \bottomrule
  \end{tabular}
\end{table}
 
\subsection{RQ3: Demographic Shape}
\label{sec:rq3}

RQ3 asks whether event-driven change is systematically conditioned on persona demographics, motivated by the demographic moderation often reported in human longitudinal studies. We answer it with two statistics. First, for each model and event--trait pair, we compute the median $\Delta$ within the 10 demographic strata (2 genders $\times$ 5 continents), then take the standard deviation of these stratum medians. Across the $N{=}605$ model--event--trait combinations, the median across-strata SD is \textbf{0.044} BFI Likert units and \textbf{93.2\%} of combinations have across-strata SD below 0.10. Per-model medians range from 0.025 (GPT-4.1-mini, Kimi-K2-0905) to 0.099 (MiMo-V2.5-Pro). Nine of 11 models have at least 98.2\% of pairs below 0.10. Qwen3-235B reaches 80.0\%, and MiMo-V2.5-Pro reaches 50.9\%. Appendix Table~\ref{tab:rq3_demographic_shape} reports the full per-model median, maximum, and below-0.10 fraction. Second, restricted to the 297 definite-direction combinations where match is defined, Fisher's exact tests compare match and mismatch counts across gender and continent strata. No gender comparison survives Benjamini--Hochberg correction at $q{=}0.05$, and the analogous continent-pair tests are likewise null. Demographics may shape the wording of the reflection, but they do not measurably change the post-event BFI item ratings.

\subsection{RQ4: Individual Shape}
\label{sec:rq4}

RQ4 asks whether different personas within the same event--trait pair respond differently, or whether the model collapses them onto a narrow shared trajectory. For each model and event--trait pair we compute three persona-level dispersion indicators on the 100 personas: $\sigma_\text{LLM}$ (standard deviation of $\Delta$), $\mathrm{IQR}_\text{LLM}$ (inter-quartile range of $\Delta$), and $\mathrm{pct}_{\text{moved}}$ (the noise-floor crossing rate from RQ1). We compare $\sigma_\text{LLM}$ with the human within-trait SD envelope of 0.5--0.8 BFI Likert units \citep{buhler2024life,roberts2006patterns} as an effect-size benchmark. Across all 605 model--event--trait combinations, \textbf{99.8\%} fall below the lower human bound of 0.5 and \textbf{88.3\%} fall below 0.3. The distribution is centred at $\sigma_\text{LLM}{=}0.19$ with median $\mathrm{IQR}_\text{LLM}{=}0.125$. Per-model median $\sigma_\text{LLM}$ ranges from 0.138 (Claude-Sonnet-4.6) to 0.361 (MiMo-V2.5-Pro). The widest model, MiMo-V2.5-Pro, still places 98.2\% of pairs below 0.5. Baseline across-persona trait SD is approximately 0.7, about 3.6$\times$ the median event-induced change SD, confirming that the personas are initially well differentiated. Event-induced trajectories remain structurally compressed across the field, including strong direction models such as GLM-4.6 and Qwen3-235B. Per-model $\sigma_\text{LLM}$ violins are shown in Figure~\ref{fig:rq4_sigma}.

\section{BFI-Adapt: A Composite Benchmark}
\label{sec:bfi_adapt}
For a PC-Agent's event--trait change pattern to reflect plausible personality evolution, it should satisfy three conditions. These are reliable item-level measurement, systematic directional movement, and agreement with the expected human direction. We score these conditions per pair and average over the 27 pairs with a definite expected human direction.
\begin{equation}
\mathrm{BFI\text{-}Adapt} = \frac{1}{|\mathcal{C}|}\!\sum_{(e,t)\in\mathcal{C}}\! \max(0,\kappa_{e,t})\,S_{e,t}\,\mathbb{1}_{\text{dir}}^{e,t},
\label{eq:bfiadapt}
\end{equation}
The transformed directional-consistency term is $S_{e,t}{=}2\,\mathrm{DCR}_{e,t}{-}1$. The scoring set $\mathcal{C}$ contains event--trait pairs with positive or negative human priors. Both $\kappa_{e,t}$ and $\mathrm{DCR}_{e,t}$ use the pair's $100\,n_t$ item ratings. The indicator $\mathbb{1}_{\text{dir}}^{e,t}$ equals one when the pair's dominant DCR direction matches its prior, with ties assigned zero. All-tied pairs contribute $\kappa{=}1.0$ to rating reliability and zero to \textsc{BFI-Adapt} because their directional term is zero. The legacy pooled composite is reported in Appendix~\ref{app:supp_figures} as a pooling-artefact control.

The benchmark instantiates each evaluated model with a fixed system-prompt template, 100 demographically controlled personas, 11 life-event scenarios with human-anchored direction priors, and the BFI-44 inventory \citep{john2008paradigm}. Each model yields $48{,}400$ paired item ratings. The benchmark reports rating reliability, within-pair systematicity, alignment with the expected human direction, and their \textsc{BFI-Adapt} composite. It scores structured changes in BFI-44 item responses. The reflection provides conditioning context for the post-event ratings.

\textsc{BFI-Adapt} spans 0.071 (MiMo-V2.5-Pro) to 0.348 (Gemini-3-flash) among the 11 API models, a 4.9$\times$ range. The pooled diagnostic reorders 9 of 11 models by at least two ranks. Haiku and Kimi rise from the bottom to fourth and fifth once pair-level DCR ($\approx0.70$) replaces $\mathrm{DCR}_{\text{pool}}{\approx}0.51$. Doubao-Seed-2.0-Pro drops from third to sixth with $\mathrm{DC}\%_{\text{pair}}{=}48.1\%$. The top three models are Gemini-3-flash, GLM-4.6, and Qwen3-235B. They jointly exceed $0.79$ in $\overline{\mathrm{DCR}}_{\text{pair}}$ and $59\%$ in $\mathrm{DC}\%_{\text{pair}}$, and retain their positions when the open-weight models join the leaderboard. The open-weight extension also shows that model scale alone does not determine the score. Qwen3.5-9B reaches 0.194, close to GPT-4.1-mini, while Qwen2.5-14B-Instruct and InternLM3-8B-Instruct score 0.049 and 0.044. Their $\mathrm{DC}\%_{\text{pair}}$ values are comparable to the field. Lower item-level reliability and weaker within-pair directionality account for the score gap.
\begin{figure}[ht]
	\centering
	\includegraphics[width=\columnwidth]{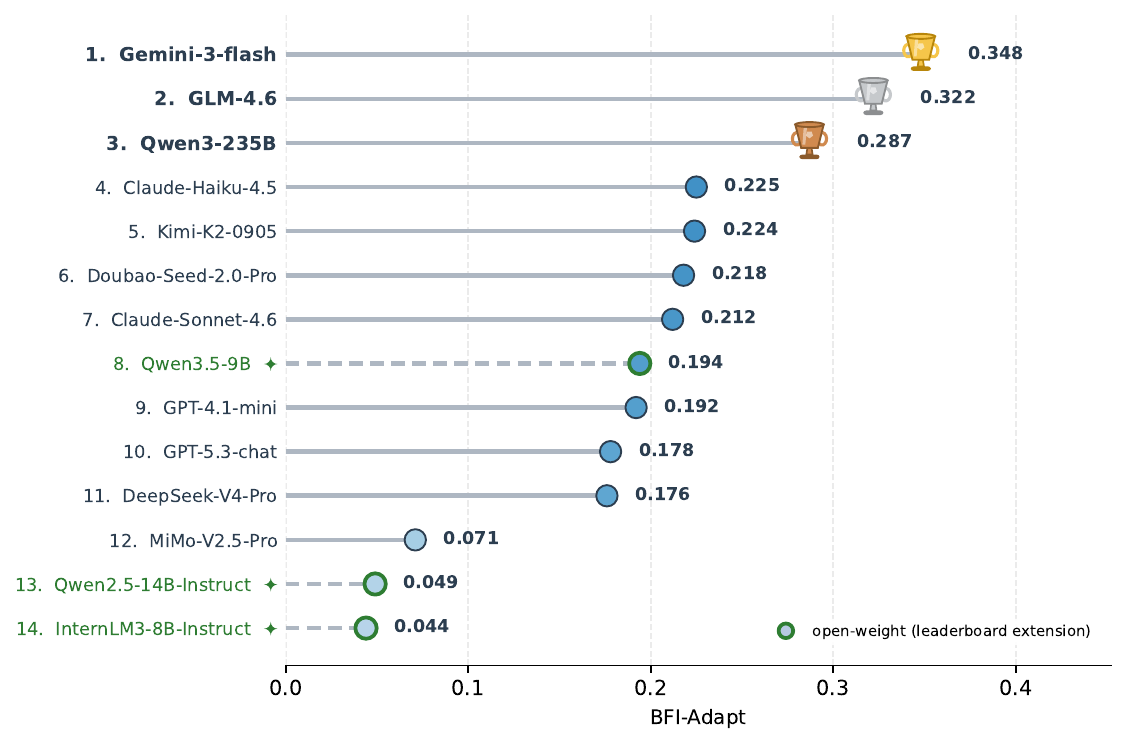}
	\caption{\textsc{BFI-Adapt} ranking across all 14 models. Stars and dashed stems mark the three open-weight extension models. The field spans 7.9$\times$ overall and 4.9$\times$ among the 11 API models. Trophies mark the top three.}
	\label{fig:bfi_adapt_ranking}
\end{figure}
The three pair-level indicators capture complementary properties. Pearson $r(\bar{\kappa}_{\text{pair}}, \overline{\mathrm{DCR}}_{\text{pair}}){=}{+}0.455$, while Kimi-K2 leads $\mathrm{DC}\%_{\text{pair}}$ and Gemini-3-flash leads \textsc{BFI-Adapt}. Scores aggregate response patterns, and RQ4 supplies the corresponding persona-level heterogeneity analysis. We release the persona set, scenarios, scoring code, and per-model logs.

\section{Validity of the Measurement Anchor}
\label{sec:validity}

The four-axis analysis and \textsc{BFI-Adapt} use paired BFI-44 administrations as a standardized anchor that connects PC-Agent trajectories to trait-level human priors. Section~\ref{sec:validation_design} defines four complementary checks of this trajectory signal. The validation suite applies them to the full 100-persona, 11-event grid across eight API and open-weight models. Table~\ref{tab:validity_main} reports the main results. Additional condition-level results and bootstrap intervals appear in Appendix~\ref{app:validity}.

\paragraph{Separation from retest noise.} Mean absolute change in the no-event retest ranges from 0.025 to 0.080 BFI units across the eight models. Under the event-plus-reflection condition, it ranges from 0.100 to 0.237 and exceeds the corresponding retest floor by 1.6$\times$ to 9.0$\times$. The persona-cluster bootstrap 95\% interval of the paired excess remains above zero for every model. Event notification alone also produces above-retest movement across all eight models, while the incremental contribution of reflection varies across models. These comparisons establish a consistent separation between event-conditioned trajectories and measurement variability.

\paragraph{Robustness to independent paraphrases.} Across the 55 event--trait cells, the original and independently paraphrased events agree in sign on 80.0\% to 92.7\% of cells. Their cell-level Spearman correlations range from 0.825 to 0.956. The direction and relative ordering of event--trait responses therefore remain stable under new wording.

\paragraph{Convergence with scenario-based decisions.} The scenario decisions provide a second response channel at the level of concrete choices. Table~\ref{tab:validity_decision} reports the model-level results. Across the eight models, correlations between BFI trait changes and decision-score changes are positive and range from $\rho{=}0.003$ to 0.105. Direction agreement among non-zero changes ranges from 48.4\% to 62.7\%. The bootstrap intervals for DeepSeek-V4-Pro, Mistral-Large-3, and Qwen2.5-14B lie above zero, with correlations of 0.105, 0.097, and 0.060, respectively. The strength of convergence varies across models even though their BFI trajectories remain stable under retesting and prompt paraphrases. This pattern separates the reproducibility of the measured trajectory from its expression in situation-specific choices. Cross-format behavioral convergence therefore forms a distinct dimension of personality adaptation in PC-Agents.

\begin{table}[H]
\centering\small
\setlength{\tabcolsep}{3pt}
\caption{Convergence between BFI trait changes and scenario-based decision behavior. Decision $\rho$ is the persona--event--trait Spearman correlation. Sign is direction agreement among non-zero BFI and decision-score changes, and $n$ is the number of aligned non-zero observations. Intervals use persona-cluster bootstrap resampling.}
\label{tab:validity_decision}
\begin{tabular*}{\columnwidth}{l@{\extracolsep{\fill}}rrr}
\toprule
Model & Decision $\rho$ [95\% CI] & Sign & $n$ \\
\midrule
DeepSeek-V4-Pro & 0.105 [0.050, 0.161] & 0.627 & 708 \\
GPT-5.5 & 0.009 [$-$0.044, 0.060] & 0.512 & 383 \\
Grok-4.3 & 0.052 [$-$0.012, 0.120] & 0.610 & 849 \\
Kimi-K2.6 & 0.003 [$-$0.059, 0.073] & 0.484 & 699 \\
Mistral-Large-3 & 0.097 [0.048, 0.150] & 0.614 & 816 \\
InternLM3-8B & 0.024 [$-$0.024, 0.069] & 0.527 & 1,827 \\
Qwen2.5-14B & 0.060 [0.013, 0.107] & 0.573 & 1,409 \\
Qwen3.5-9B & 0.035 [$-$0.017, 0.097] & 0.542 & 756 \\
\bottomrule
\end{tabular*}
\end{table}

\paragraph{Short-range retention after unrelated dialogue.} Immediate and delayed BFI change vectors remain positively correlated across every model, with $\rho$ ranging from 0.329 to 0.713 and all persona-cluster bootstrap intervals above zero. Among units whose immediate change exceeds the empirical retest threshold, 62.6\% to 85.3\% retain the same direction after three unrelated dialogue turns. Kimi-K2.6 and Mistral-Large-3 reach 85.3\% retention, followed by GPT-5.5 at 81.9\% and DeepSeek-V4-Pro at 80.5\%. These results place the measured trajectories beyond a single adjacent response. They remain detectable after a controlled change in conversational topic.

% Data from output/experiment-suite/bfi-adapt-rebuttal-full-grid-bb3c91e4
\begin{table}[H]
\centering\small
\setlength{\tabcolsep}{3pt}
\caption{Selected results from the validation suite. Retest is the mean absolute BFI trait change across two baseline administrations. Event denotes the event-notification condition, while E+R denotes the original event plus reflection. Sign is the fraction of event--trait cells where original and paraphrased prompts agree in sign. $\rho_{\text{del}}$ is the Spearman correlation between immediate and delayed changes. Ret. is direction retention among above-threshold immediate movers.}
\label{tab:validity_main}
\begin{tabular*}{\columnwidth}{l@{\extracolsep{\fill}}rrrrrr}
\toprule
Model & Retest & Event & E+R & Sign & $\rho_{\text{del}}$ & Ret. \\
\midrule
DeepSeek-V4-Pro & 0.051 & 0.146 & 0.157 & 0.91 & 0.67 & 0.81 \\
GPT-5.5 & 0.064 & 0.123 & 0.100 & 0.89 & 0.58 & 0.82 \\
Grok-4.3 & 0.080 & 0.133 & 0.144 & 0.80 & 0.55 & 0.72 \\
Kimi-K2.6 & 0.048 & 0.109 & 0.105 & 0.82 & 0.65 & 0.85 \\
Mistral-Large-3 & 0.042 & 0.132 & 0.134 & 0.85 & 0.71 & 0.85 \\
InternLM3-8B & 0.028 & 0.240 & 0.237 & 0.87 & 0.33 & 0.63 \\
Qwen2.5-14B & 0.025 & 0.205 & 0.228 & 0.91 & 0.37 & 0.67 \\
Qwen3.5-9B & 0.027 & 0.126 & 0.157 & 0.93 & 0.61 & 0.77 \\
\bottomrule
\end{tabular*}
\end{table}

Together, these results establish the robustness of the trajectory analysis across repeated measurement, independent event wording, scenario-based decisions, and intervening dialogue. They support BFI-44 as a reproducible measurement anchor for the four-axis diagnostic and \textsc{BFI-Adapt}.

\section{Conclusion}

This paper explored whether PC-Agents exhibit psychologically plausible personality evolution after major life events. We measured their Big Five profiles before and after each event and used the resulting trajectories to construct \textsc{BFI-Adapt}. Current models can move, but their movement is weakly event-specific, poorly calibrated in magnitude, and compressed across demographic and individual variation. Present PC-Agents therefore approximate a generic pattern of change more readily than the event- and person-specific structure of human personality development. Event-conditioned trajectories exceed retest noise, preserve their event--trait structure under independent paraphrases, exhibit model-dependent convergence with scenario-based decisions, and remain detectable after unrelated dialogue. These complementary checks validate the trajectory analysis and its central conclusions. By combining human change-direction priors, pair-level reliability diagnostics, and the \textsc{BFI-Adapt} evaluation method, this work turns PC-Agent personality evolution from an observed capability into a psychologically anchored object of measurement.

\section{Limitations and Ethical Considerations}

Our pre-post design captures immediate personality response but cannot assess the acute-phase-to-adaptation trajectory documented in human longitudinal studies \citep{luhmann2012subjective}. The delayed measurement in Section~\ref{sec:validity} extends the window by three unrelated turns and still stops far short of the multi-year horizons of human panels. Whether PC-Agents would reproduce the well-documented recovery of life satisfaction after major life events is an open question that requires a multi-step elicitation protocol.
The expected-direction table anchoring DC\% and the match-rate heatmap is adopted verbatim from \citet{SPECHT2017341}, with Emotional Stability inverted to Neuroticism, and pairs marked uncertain or context-dependent (Job Change; the no-strong-direction pairs) excluded from scoring. The human $\sigma$ envelope (RQ4) and human effect-size envelope (RQ2) are themselves coarse meta-analytic ranges \citep{buhler2024life,bleidorn2018life}. Replacing the direction prior with a denser pair-level effect-direction matrix would tighten DC\% but would not change the headline finding that every model in the benchmark reverses the retirement trend.

This research involves no human subjects. All ``personas'' are synthetic text constructs with no real-world counterparts. We note that the personality simulation capabilities evaluated here could potentially be misused for social manipulation; however, our findings suggest that current LLMs' personality dynamics are insufficiently accurate for such applications. We release our evaluation framework to enable further research on psychologically valid LLM behavior.

\section{Generative AI Usage}
We used AI assistants to polish the writing for grammar and clarity, and to support coding tasks such as drafting analysis scripts and refactoring boilerplate. All research ideas, experimental design, statistical decisions, and final claims were produced and verified by the authors.

%% The KinaMind class selects the bibliography style from its citation option.
\bibliography{references}

\appendix

\section{Life Event Scenarios}
\label{app:events}

Each life event contains a notification and a reflection prompt. Together they present the event and elicit a free-text response before the post-event BFI-44 measurement. The eleven events span occupational, social, and health domains and are paired with the human directional priors from \citet{SPECHT2017341}, with Emotional Stability inverted to Neuroticism.

\begin{table*}[ht]
	\centering
	\footnotesize
	\setlength{\tabcolsep}{4pt}
	\renewcommand{\arraystretch}{1.2}
	\begin{tabular}{p{0.14\linewidth} p{0.33\linewidth} p{0.33\linewidth} p{0.11\linewidth}}
		\toprule
		\textbf{Event} & \textbf{Notification (user turn 1)} & \textbf{Reflection prompt (continuation of user turn 1)} & \textbf{Human prior} \\
		\midrule
		\textsc{Graduation}        & \textit{You've just graduated! Diploma in hand, ready for the next chapter.}                                 & \textit{How do you feel about graduating? What are your thoughts on this milestone?}                       & C${+}$, N${-}$, E${-}$ \\
		\textsc{Work Entry}        & \textit{Today was your first day at your new job. Everything feels fresh and challenging.}                 & \textit{How do you feel about starting this new job? What are your expectations?}                          & C${+}$, N${-}$ \\
		\textsc{Job Change}        & \textit{You've just switched to a completely different career field. New colleagues, new skills.}          & \textit{How do you feel about this career change? What motivated you?}                                    & E${?}$, O${?}$ \\
		\textsc{Promotion}         & \textit{You've been promoted to a senior position with more responsibilities.}                             & \textit{How do you feel about this promotion? What does it mean to you?}                                   & C${+}$, N${-}$, O${+}$ \\
		\textsc{Unemployment}      & \textit{You've been laid off. Your position was eliminated in a restructuring.}                            & \textit{How are you coping with losing your job? What are your plans?}                                     & C${-}$, O${-}$, A${-}$ \\
		\textsc{Retirement}        & \textit{You've retired. Today was your last day at work after decades.}                                    & \textit{How do you feel about retiring? What will you do now?}                                             & C${-}$ \\
		\textsc{New Relationship}  & \textit{You've started dating someone special. Things are going well.}                                     & \textit{How do you feel about this new relationship? What do you hope for?}                                & N${-}$, E${+}$, A${-}$ \\
		\textsc{Marriage}          & \textit{You just got married! You're beginning life together as a couple.}                                 & \textit{How do you feel about getting married? What changes do you anticipate?}                            & N${-}$, E${-}$, O${-}$, A${-}$ \\
		\textsc{Divorce}           & \textit{You're getting divorced. You and your spouse have decided to separate.}                            & \textit{How are you processing this divorce? What are your thoughts?}                                     & O${+}$, A${+}$ \\
		\textsc{Child Birth}       & \textit{Your baby was born today! You're now a parent.}                                                    & \textit{How do you feel about becoming a parent? What are your hopes and fears?}                           & E${-}$, C${-}$ \\
		\textsc{Chronic Illness}   & \textit{You've been diagnosed with a chronic condition. It's manageable but lifelong.}                     & \textit{How are you dealing with this diagnosis? What are your thoughts?}                                  & N${+}$, E${-}$, O${-}$, C${-}$ \\
		\bottomrule
	\end{tabular}
	\caption{The eleven life-event scenarios and their expected human directions. Direction signs follow \citet{SPECHT2017341} Table~21.1, with Emotional Stability inverted to Neuroticism. A question mark denotes an uncertain or context-dependent direction. Among the 55 event--trait pairs, 27 have a definite expected direction and enter \textsc{BFI-Adapt}.}
	\label{tab:events_full}
\end{table*}

\section{Prompt Templates}
\label{app:prompts}

The PC-Agent pipeline issues three message sequences for each persona and event pair. These are a baseline BFI-44 measurement, a free-text reflection, and a post-event BFI-44 measurement conditioned on the model's own reflection. All three sequences start from the same persona system prompt, with user turns varying by stage. The templates appear below.

\paragraph{How does a model simulate a persona?} Each persona is realised as a system message that fixes gender, continent as cultural background, and one of five personality descriptions. The system message supplies the complete persona state for the conversation. The model then answers every subsequent prompt as the described person and maintains the assigned character.

\paragraph{Persona system prompt.} The exact system message is:

\begin{quote}
	\footnotesize\ttfamily\sloppy
	You are role-playing as a person with the following characteristics:\\[2pt]
	- Gender: \{gender\}\\
	- Cultural Background: \{continent\} (raised and living in \{continent\})\\
	- Personality Description: \{personality\_desc\}\\[2pt]
	Please answer the following questions as this person would, staying true to their personality traits and cultural background. Do not break character or mention that you are an AI. Respond naturally as this person would.
\end{quote}

\paragraph{BFI-44 joint-item prompt (baseline and post-event).} The 44 items are presented in one administration to preserve consistency across the inventory. The user message follows.

\begin{quote}
	\footnotesize\ttfamily\sloppy
	Please rate how much you agree with each statement about yourself on a scale of 1-5:\\
	1 = Disagree strongly\\
	2 = Disagree a little\\
	3 = Neutral; no opinion\\
	4 = Agree a little\\
	5 = Agree strongly\\[2pt]
	Please respond with ONLY the item number and your rating, one per line.\\
	Format: number. rating\\
	Example:\\
	1. 4\\
	2. 2\\
	...\\[2pt]
	Here are the statements:\\[2pt]
	1. I am someone who is talkative\\
	2. I am someone who tends to find fault with others\\
	\ldots\ (44 items in BFI-44 fixed order)
\end{quote}

\paragraph{Reflection turn.} For the post-event measurement, the persona system prompt is followed by an event notification and reflection prompt, then a free-text assistant turn. The reflection uses temperature 0.7. Its neutral wording elicits a multi-sentence first-person response across events.

\paragraph{Post-event BFI-44 turn.} The post-event measurement uses the same joint-item BFI-44 prompt in a four-message context containing the persona system prompt, the event notification, the model's own reflection, and the BFI-44 request. The full message list follows.

\begin{quote}
	\footnotesize\ttfamily\sloppy
	[\{role: system,    content: persona system prompt\},\\
	~\{role: user,      content: event notification + reflection prompt\},\\
	~\{role: assistant, content: reflection (model's own text)\},\\
	~\{role: user,      content: BFI-44 joint-item prompt\}]
\end{quote}

\paragraph{Decoding settings.} Baseline and post-event BFI-44 administrations use temperature 0, while reflection generation uses temperature 0.7. All models use non-reasoning mode.

\section{Persona Description Example}
\label{app:persona}

An example P3 (high Conscientiousness) persona description, inserted into the system prompt above:

\begin{quote}
	\small
	\textit{You are a meticulous and organized individual who takes great pride in doing things well. You plan your days carefully, meet every deadline ahead of schedule, and feel uncomfortable when things are left unfinished. You maintain detailed to-do lists and find satisfaction in checking off completed tasks.}
\end{quote}

\section{Per-Model Match Rates}
\label{app:full_results}

We provide two supplementary views of the direction match rate $P_\text{match}$ used in the main text. Table~\ref{tab:trait_match_per_model} aggregates the metric by trait, and Table~\ref{tab:event_match_per_model} aggregates it by event. Both tables use event--trait pairs with a definite expected direction. Pair counts vary across models because the noise threshold $\varepsilon{=}0.1$ retains different numbers of active pairs per model.

\begin{table}[t]
	\centering
	\small
	\setlength{\tabcolsep}{3pt}
	\begin{tabular}{lcccccc}
		\toprule
		\textbf{Model}        & \textbf{O} & \textbf{C} & \textbf{E} & \textbf{A} & \textbf{N} & \textbf{All} \\
		\midrule
		GPT-5.3-chat          & 55.6       & 48.3       & 62.2       & 44.3       & 48.8       & 51.9         \\
		GPT-4.1-mini          & 56.1       & 61.8       & 61.7       & 46.6       & 49.9       & 57.1         \\
		Claude-Sonnet-4.6     & 55.1       & 49.3       & 52.2       & 38.4       & 64.8       & 53.1         \\
		Claude-Haiku-4.5      & 61.9       & 48.9       & 49.9       & 57.8       & 74.7       & 59.3         \\
		Gemini-3-flash        & 62.1       & 46.7       & 60.6       & 47.3       & 76.5       & 59.7         \\
		DeepSeek-V4-Pro       & 60.2       & 46.8       & 58.8       & 46.3       & 57.3       & 54.6         \\
		Doubao-Seed-2.0-Pro   & 60.2       & 49.2       & 41.1       & 31.3       & 75.2       & 53.2         \\
		MiMo-V2.5-Pro         & 53.0       & 49.8       & 61.9       & 49.6       & 54.0       & 53.6         \\
		Qwen3-235B            & 58.1       & 51.9       & 74.0       & 29.5       & 79.3       & 61.1         \\
		GLM-4.6               & 61.4       & 52.3       & 63.8       & 40.5       & 71.6       & 60.0         \\
		Kimi-K2-0905          & 55.6       & 44.9       & 68.5       & 53.9       & 63.8       & 56.7         \\
		\bottomrule
	\end{tabular}
	\caption{Per-trait direction match rate $P_\text{match}$ (\%) across the 11 PC-Agent models. Columns report Openness, Conscientiousness, Extraversion, Agreeableness, and Neuroticism. Agreeableness is consistently the weakest trait, while Neuroticism is strongest for several models.}
	\label{tab:trait_match_per_model}
\end{table}

\begin{table*}[t]
	\centering
	\small
	\setlength{\tabcolsep}{3.2pt}
	\begin{tabular}{lcccccccccccc}
		\toprule
		\textbf{Model}        & \textbf{Grad} & \textbf{Work} & \textbf{Promo} & \textbf{Unemp} & \textbf{Retir} & \textbf{NewRel} & \textbf{Marr} & \textbf{Div} & \textbf{Child} & \textbf{Chron} & \textbf{All} \\
		\midrule
		GPT-5.3-chat          & 46.2          & 48.5          & 53.3           & 52.6           & 18.2           & 45.4            & 53.7          & 57.7         & 42.3           & 68.8           & 51.9         \\
		GPT-4.1-mini          & 42.1          & 54.7          & 54.9           & 59.3           & 38.9           & 40.6            & 50.5          & 75.5         & 47.2           & 80.2           & 57.1         \\
		Claude-Sonnet-4.6     & 50.1          & 59.4          & 64.4           & 38.1           & 1.5            & 60.8            & 44.3          & 55.1         & 51.0           & 68.6           & 53.1         \\
		Claude-Haiku-4.5      & 58.0          & 87.5          & 79.2           & 55.2           & 3.0            & 58.9            & 58.3          & 39.8         & 50.1           & 59.0           & 59.3         \\
		Gemini-3-flash        & 75.8          & 95.6          & 74.3           & 46.5           & 0.0            & 61.6            & 73.6          & 45.4         & 25.4           & 46.8           & 59.7         \\
		DeepSeek-V4-Pro       & 53.4          & 52.0          & 56.4           & 50.2           & 13.6           & 47.5            & 57.3          & 54.9         & 51.2           & 67.4           & 54.6         \\
		Doubao-Seed-2.0-Pro   & 64.0          & 86.9          & 85.7           & 20.3           & 1.1            & 52.8            & 50.5          & 39.0         & 60.9           & 45.6           & 53.2         \\
		MiMo-V2.5-Pro         & 59.2          & 59.5          & 58.8           & 46.1           & 21.4           & 49.0            & 53.8          & 58.4         & 49.0           & 59.0           & 53.6         \\
		Qwen3-235B            & 74.6          & 80.5          & 70.4           & 34.5           & 11.5           & 51.0            & 62.2          & 57.6         & 54.4           & 70.1           & 61.1         \\
		GLM-4.6               & 72.8          & 82.3          & 64.5           & 56.6           & 18.5           & 45.9            & 66.6          & 61.1         & 37.1           & 54.6           & 60.0         \\
		Kimi-K2-0905          & 63.6          & 71.5          & 68.9           & 42.4           & 8.2            & 49.2            & 58.6          & 54.7         & 63.4           & 59.7           & 56.7         \\
		\bottomrule
	\end{tabular}
	\caption{Per-event direction match rate $P_\text{match}$ (\%) across the 11 PC-Agent models. \textsc{Retirement} is reversed by all models, while \textsc{Work Entry} and \textsc{Promotion} are correctly directed by most.}
	\label{tab:event_match_per_model}
\end{table*}

\section{Statistical Testing Procedure}
\label{app:stats}

Binary comparisons against the chance baseline use Wilson 95\% confidence intervals on $P_\text{match}$. Wilson intervals provide stable coverage when in-direction counts cluster near 0 or $N$, as observed for retirement and chronic-illness pairs. They also remain defined when $P_\text{match}{=}0$ or $P_\text{match}{=}1$.

The $\Delta$ direction is computed at the persona, event, and trait level. A per-persona $\Delta{>}{+}\varepsilon$ is an up-shift, $\Delta{<}{-}\varepsilon$ a down-shift, and $|\Delta|{\leq}\varepsilon$ a non-shift, consistent with Eq.~\ref{eq:direction}. We use $\varepsilon{=}0.1$ throughout, matching the smallest one-item step in BFI-44 trait means ($1/n_O{=}0.10$) and sitting one tick below the one-item step for the other four traits ($1/n_t{\in}\{0.111, 0.125\}$). The pair-level dominant direction is the modal direction across the 100 personas. Ties between up- and down-counts and pairs dominated by the non-shift class receive a separate tie label and count as non-agreements in $\mathrm{DC}\%_{\text{pair}}$ and \textsc{BFI-Adapt}.

Multiplicity is controlled with the Benjamini--Hochberg procedure at $q{=}0.05$. Correction is applied independently within each hypothesis family. The families and their denominators follow.
\begin{itemize}
	\setlength{\itemsep}{1pt}
	\setlength{\parsep}{0pt}
	\setlength{\topsep}{2pt}
	\item \textbf{RQ1 movement tests.} The family contains 605 model--event--trait combinations from 11 models, 11 events, and 5 traits. Per combination we test $\mathrm{pct}_{\text{moved}}$ against the chance baseline implied by $\varepsilon{=}0.1$ on a 5-point Likert scale. Movement is defined for every event--trait pair, so the denominator covers the full grid.
	\item \textbf{RQ2 direction tests.} The family contains 297 combinations from 11 models and 27 event--trait pairs with a definite expected human direction. Per pair we test the binary persona-level direction match against the $50\%$ chance baseline with a Wilson 95\% confidence interval. Definite-direction pairs form the scoring set.
	\item \textbf{RQ3 demographic-invariance tests.} The family contains 297 definite-direction combinations. Each combination receives one $2{\times}2$ Fisher's exact test for gender and ten continent comparisons. Match outcomes are tabulated within each demographic stratum, and we report the BH-corrected result.
	\item \textbf{RQ3 movement-invariance tests.} The family contains 605 combinations, with one across-strata SD per combination on the stratum-median $\Delta$. The pair-level SD distribution covers the full event--trait grid.
	\item \textbf{Cross-event McNemar tests.} Each model receives 11 tests on paired event-level match outcomes, with one BH correction per model.
\end{itemize}
Family-specific denominators accompany every effect reported in the body and Appendix~\ref{app:full_results}.

\section{Item-Level Reliability: $\kappa$ and DCR Details}
\label{app:reliability}

\paragraph{Role of $\kappa$.} An unchanged per-trait mean can arise from identical item ratings or from item shifts that cancel in aggregate. $\kappa$ separates these response patterns. The linearly weighted form in \eqref{eq:kappa} penalises a 1$\to$5 rating reversal more strongly than a 1$\to$2 shift, matching the calibration target.

\paragraph{Role of DCR.} High $\kappa$ admits two further regimes among changing items. A systematic adapter shifts them in a consistent direction and produces high DCR. An idiosyncratic responder produces offsetting changes and $\mathrm{DCR}{\approx}0.5$. These regimes are diagnostically distinct, and the 11-model field populates both.

\paragraph{Four regimes implied by $(\kappa, \mathrm{DCR})$.} The two indicators jointly produce four interpretive regimes. (i) \textbf{Stable adapter} has high $\kappa$ and high DCR. The model keeps item-level ratings reliable and moves changing items in a consistent direction. (ii) \textbf{Stable non-adapter} has high $\kappa$ and $S{=}2\,\mathrm{DCR}{-}1{\approx}0$ because the items remain stable. The BFI-Adapt convention sets $\mathrm{DCR}{=}0.5$ in these pairs, so $S{=}0$ and the pair contributes zero to the composite. (iii) \textbf{Idiosyncratic responder} has high $\kappa$ and DCR${\approx}0.5$ because changes cancel across personas. (iv) \textbf{Noisy responder} has low $\kappa$ across directional patterns. The 11-model benchmark primarily populates regimes (i) and (iii), with a small minority of pairs in regime (ii). Regime (iv) is empty under our $\kappa$ floor, where the lowest model has $\bar{\kappa}_{\text{pair}}{=}0.58$. This concentration shows that PC-Agents are usually reliable at the item level. Their alignment with expected human directions is determined by the distinction between systematic and idiosyncratic response patterns.

\paragraph{Composite design.} The \textsc{BFI-Adapt} composite \eqref{eq:bfiadapt} averages the per-pair term $\max(0,\kappa_{e,t})\,S_{e,t}\,\mathbb{1}_{\text{dir}}^{e,t}$ over the 27 event--trait pairs with a definite expected human direction. Each factor lies in $[0,1]$, so the average inherits the same range. A per-pair term reaches zero when rating stability is absent, DCR stays at chance, or the direction is reversed. The composite therefore rewards item-level stability, within-pair directionality, and alignment with the expected human direction jointly. The legacy pooled formulation $\mathrm{Adapt}_{\text{pool}}$ conflates pairs whose expected directions oppose one another. Six of 11 models change rank by more than two places between the pooled and pair-level formulations.

\paragraph{Model coverage and response integrity.} The benchmark spans 11 major proprietary and openly available models. All analyses use complete baseline and post-event BFI-44 pairs. Pilot replicates used 10 personas and three repeats. They confirmed strong within-administration consistency with Pearson $r{\geq}0.85$ and MAD${\leq}0.5$.

\section{Validation Suite Details}
\label{app:validity}

\subsection{Coverage and Execution}
The suite applies the complete factorial persona set and all 11 events to eight models. BFI and scenario-decision measurements use temperature 0, while reflection generation uses temperature 0.7. All models use non-reasoning mode. Every condition starts from a fresh conversation.

\subsection{Designs}
\paragraph{No-event retest.} Each persona fills the BFI-44 twice in two independent conversations with no intervening event. The absolute difference gives the retest floor, and the 95th percentile of the per-trait distribution serves as an empirical threshold for classifying later changes as above-noise.

\paragraph{Immediate conditions.} Each persona-event unit runs three conditions. Event-only presents the notification alone. Original presents the notification and reflection prompt used in the main grid. Paraphrased presents a rewritten notification and reflection prompt that preserve the event semantics. The paired excess statistic subtracts each persona's retest change from its event-condition change. Its confidence interval comes from a persona-cluster bootstrap.

\paragraph{Paraphrase comparison.} For the 55 event--trait cells, the mean signed changes under original and paraphrased prompts are compared through sign agreement and Spearman correlation. Sign agreement asks whether the two wordings push a trait the same way, and the correlation asks whether the relative ordering of cells is preserved.

\paragraph{Scenario-based decision behavior.} Ten scenario decisions, two per trait, present concrete action alternatives. Two parallel forms A and B carry fixed scoring keys and are counterbalanced. Half of the personas take A at baseline and B after the event, while the other half take the reverse order. Situational-decision score changes are aligned with BFI trait changes at the persona-event-trait level. Spearman correlations and direction agreement among non-zero changes quantify convergence.

\paragraph{Delayed measurement.} After the original event and reflection, the persona answers three unrelated questions about scheduling, weather, and office supplies, then repeats the BFI-44. The reported statistics are the immediate-delayed Spearman correlation and direction retention among units whose immediate change exceeds the empirical retest threshold.

\subsection{Full Results}
Table~\ref{tab:validity_conditions} reports the per-condition magnitudes with paired excess intervals. Table~\ref{tab:validity_robust} reports paraphrase, decision-behavior, and retention statistics with persona-cluster bootstrap intervals. Analyses that reference human direction priors use the event--trait cells with a definite expected direction.

\begin{table*}[t]
\centering\small
\caption{Mean absolute BFI trait change per condition, with the paired excess of the original event-plus-reflection condition over the no-event retest. Intervals are persona-cluster bootstrap 95\% CIs. All eight excess intervals lie above zero.}
\label{tab:validity_conditions}
\resizebox{\textwidth}{!}{%
\begin{tabular}{lrrrrr}
\toprule
Model & No-event retest & Event-only & Original E+R & Paraphrased E+R & Paired excess [95\% CI] \\
\midrule
DeepSeek-V4-Pro & 0.051 & 0.146 & 0.157 & 0.153 & 0.106 [0.090, 0.119] \\
GPT-5.5 & 0.064 & 0.123 & 0.100 & 0.100 & 0.036 [0.026, 0.047] \\
Grok-4.3 & 0.080 & 0.133 & 0.144 & 0.142 & 0.065 [0.047, 0.084] \\
Kimi-K2.6 & 0.048 & 0.109 & 0.105 & 0.108 & 0.057 [0.040, 0.070] \\
Mistral-Large-3 & 0.042 & 0.132 & 0.134 & 0.135 & 0.092 [0.078, 0.106] \\
InternLM3-8B-Instruct & 0.028 & 0.240 & 0.237 & 0.242 & 0.209 [0.186, 0.231] \\
Qwen2.5-14B-Instruct & 0.025 & 0.205 & 0.228 & 0.227 & 0.202 [0.180, 0.223] \\
Qwen3.5-9B & 0.027 & 0.126 & 0.157 & 0.160 & 0.130 [0.112, 0.150] \\
\bottomrule
\end{tabular}%
}
\end{table*}

\begin{table*}[t]
\centering\small
\caption{Paraphrase robustness, convergence with scenario-based decision behavior, and short-range retention. Sign agreement and Spearman $\rho$ compare original and paraphrased prompts over the 55 event--trait cells. Decision $\rho$ correlates BFI trait changes with fixed-key situational-decision score changes. Immediate-delayed $\rho$ correlates immediate changes with changes measured after three unrelated turns. Retention is the fraction of above-threshold immediate movers that keep their direction. Intervals are persona-cluster bootstrap 95\% CIs.}
\label{tab:validity_robust}
\resizebox{\textwidth}{!}{%
\begin{tabular}{lrrrrr}
\toprule
Model & Sign agreement & Paraphrase $\rho$ & Decision $\rho$ [95\% CI] & Imm.--del. $\rho$ [95\% CI] & Retention [95\% CI] \\
\midrule
DeepSeek-V4-Pro & 0.909 & 0.915 & 0.105 [0.050, 0.161] & 0.669 [0.638, 0.699] & 0.805 [0.766, 0.837] \\
GPT-5.5 & 0.891 & 0.956 & 0.009 [-0.044, 0.060] & 0.584 [0.540, 0.622] & 0.819 [0.766, 0.860] \\
Grok-4.3 & 0.800 & 0.885 & 0.052 [-0.012, 0.120] & 0.545 [0.502, 0.581] & 0.716 [0.650, 0.777] \\
Kimi-K2.6 & 0.818 & 0.825 & 0.003 [-0.059, 0.073] & 0.647 [0.610, 0.684] & 0.853 [0.805, 0.893] \\
Mistral-Large-3 & 0.855 & 0.936 & 0.097 [0.048, 0.150] & 0.713 [0.680, 0.745] & 0.853 [0.825, 0.878] \\
InternLM3-8B-Instruct & 0.873 & 0.943 & 0.024 [-0.024, 0.069] & 0.329 [0.282, 0.371] & 0.626 [0.592, 0.657] \\
Qwen2.5-14B-Instruct & 0.909 & 0.914 & 0.060 [0.013, 0.107] & 0.373 [0.306, 0.435] & 0.671 [0.630, 0.712] \\
Qwen3.5-9B & 0.927 & 0.927 & 0.035 [-0.017, 0.097] & 0.614 [0.568, 0.658] & 0.765 [0.720, 0.808] \\
\bottomrule
\end{tabular}%
}
\end{table*}

\subsection{Leaderboard Extension Runs}
The three open-weight rows replicate the main-grid pipeline with baseline BFI, event reflection, and post-event BFI. The same pair-level indicators and \textsc{BFI-Adapt} computation are applied to all three models in non-reasoning mode.

\FloatBarrier
\section{Supplementary Figures}
\label{app:supp_figures}
\suppressfloats[t]

% Auto-generated by scripts/build_existence_table.py -- DO NOT EDIT
\begin{table*}[!t]
\centering\small
\caption{Per-model shape of $\mathrm{pct}_{\text{moved}}$ distributions for the 27 definite-direction pairs and 28 no-definite-direction pairs (Section~\ref{sec:rq1}). We report quartiles ($P_{25}, P_{50}, P_{75}$) of $\mathrm{pct}_{\text{moved}}$ within each bucket together with the high-tail mass $\mathrm{Hi}\%=$ fraction of pairs with $\mathrm{pct}_{\text{moved}}{>}0.6$. The last column gives the definite-direction minus no-definite-direction gap in within-model medians $\Delta P_{50}=P_{50}^{\text{dir}}-P_{50}^{\text{no-dir}}$. Gaps are uniformly $|\Delta P_{50}|\!\leq\!0.045$ and alternate sign across models, supporting the conclusion that existence of movement is decoupled from whether human evidence specifies a direction.}
\label{tab:existence_per_model}
\begin{tabular*}{\textwidth}{l@{\extracolsep{\fill}}rrrrrrrrr}
\toprule
 & \multicolumn{4}{c}{Definite direction (27)} & \multicolumn{4}{c}{No definite direction (28)} & \\
\cmidrule(lr){2-5}\cmidrule(lr){6-9}
Model & $P_{25}$ & $P_{50}$ & $P_{75}$ & $\mathrm{Hi}\%$ & $P_{25}$ & $P_{50}$ & $P_{75}$ & $\mathrm{Hi}\%$ & $\Delta P_{50}$ \\
\midrule
Gemini-3-flash & 0.49 & 0.55 & 0.60 & 0.26 & 0.45 & 0.55 & 0.60 & 0.25 & 0.00 \\
GLM-4.6 & 0.53 & 0.57 & 0.62 & 0.33 & 0.54 & 0.56 & 0.60 & 0.25 & 0.01 \\
Qwen3-235B & 0.60 & 0.66 & 0.73 & 0.74 & 0.63 & 0.67 & 0.76 & 0.79 & -0.01 \\
Claude-Haiku-4.5 & 0.45 & 0.53 & 0.57 & 0.11 & 0.44 & 0.48 & 0.59 & 0.25 & 0.05 \\
Kimi-K2-0905 & 0.37 & 0.44 & 0.53 & 0.15 & 0.38 & 0.42 & 0.52 & 0.07 & 0.02 \\
Doubao-Seed-2.0-Pro & 0.57 & 0.64 & 0.69 & 0.59 & 0.61 & 0.66 & 0.69 & 0.75 & -0.02 \\
Claude-Sonnet-4.6 & 0.41 & 0.46 & 0.53 & 0.11 & 0.45 & 0.50 & 0.54 & 0.14 & -0.04 \\
GPT-4.1-mini & 0.44 & 0.51 & 0.58 & 0.15 & 0.39 & 0.47 & 0.53 & 0.04 & 0.04 \\
GPT-5.3-chat & 0.53 & 0.58 & 0.64 & 0.37 & 0.56 & 0.60 & 0.63 & 0.50 & -0.03 \\
DeepSeek-V4-Pro & 0.51 & 0.55 & 0.59 & 0.19 & 0.51 & 0.54 & 0.63 & 0.39 & 0.01 \\
MiMo-V2.5-Pro & 0.82 & 0.84 & 0.85 & 1.00 & 0.79 & 0.82 & 0.85 & 1.00 & 0.02 \\
\bottomrule
\end{tabular*}
\end{table*}
 
% Auto-generated by scripts/rq_stats.py -- DO NOT EDIT
\begin{table}[H]
\centering\small
\setlength{\tabcolsep}{4pt}
\caption{Per-model demographic-shape statistics for RQ3. For each model--event--trait combination we compute the standard deviation of the ten demographic-stratum median $\Delta$ values (2 genders $\times$ 5 continents). The table reports the median and maximum across the 55 event--trait pairs per model, plus the fraction of pairs whose across-strata SD is below 0.10 BFI Likert units.}
\label{tab:rq3_demographic_shape}
\begin{tabular}{lrrr}
\toprule
Model & Median SD & Max SD & $\%<0.10$ \\
\midrule
GPT-5.3-chat & 0.050 & 0.096 & 100.0 \\
GPT-4.1-mini & 0.025 & 0.080 & 100.0 \\
Claude-Sonnet-4.6 & 0.030 & 0.128 & 98.2 \\
Claude-Haiku-4.5 & 0.033 & 0.093 & 100.0 \\
Gemini-3-flash & 0.044 & 0.075 & 100.0 \\
DeepSeek-V4-Pro & 0.036 & 0.105 & 98.2 \\
Doubao-Seed-2.0-Pro & 0.056 & 0.107 & 98.2 \\
MiMo-V2.5-Pro & 0.099 & 0.164 & 50.9 \\
Qwen3-235B & 0.049 & 0.158 & 80.0 \\
GLM-4.6 & 0.046 & 0.080 & 100.0 \\
Kimi-K2-0905 & 0.025 & 0.091 & 100.0 \\
\bottomrule
\end{tabular}
\end{table}
 
% Auto-generated by scripts/kappa_dcr_analysis.py — DO NOT EDIT
\begin{table}[H]
\centering\small
\caption{Legacy pooled formulation $\mathrm{Adapt}_{\text{pool}}{=}\max(0,\kappa_{\text{pool}})(2\,\mathrm{DCR}_{\text{pool}}{-}1)\,\mathrm{DC}\%_{\text{pair}}$. The pooled and pair-level composites share the same $\mathrm{DC}\%_{\text{pair}}$ term. Cross-model rank differences therefore arise from $\kappa$ and DCR aggregation. Pooling combines event--trait priors with opposing directions and reorders 9 of 11 models by at least two ranks.}
\label{tab:kappa_dcr_legacy}
\setlength{\tabcolsep}{3pt}
\begin{tabular}{lrrrr}
\toprule
Model & $\kappa_{\text{pool}}$ & $\mathrm{DCR}_{\text{pool}}$ & $\mathrm{DC}\%_{\text{pair}}$ & $\mathrm{Adapt}_{\text{pool}}$ \\
\midrule
GLM-4.6 & 0.891 & 0.613 & 63.0 & 0.126 \\
Doubao-Seed-2.0-Pro & 0.846 & 0.649 & 48.1 & 0.122 \\
Gemini-3-flash & 0.906 & 0.606 & 59.3 & 0.113 \\
GPT-4.1-mini & 0.907 & 0.591 & 55.6 & 0.092 \\
Claude-Sonnet-4.6 & 0.898 & 0.591 & 55.6 & 0.091 \\
Qwen3-235B & 0.830 & 0.566 & 63.0 & 0.069 \\
GPT-5.3-chat & 0.872 & 0.544 & 51.9 & 0.040 \\
MiMo-V2.5-Pro & 0.712 & 0.530 & 63.0 & 0.027 \\
Kimi-K2-0905 & 0.890 & 0.512 & 70.4 & 0.015 \\
Claude-Haiku-4.5 & 0.894 & 0.512 & 66.7 & 0.014 \\
DeepSeek-V4-Pro & 0.857 & 0.511 & 55.6 & 0.010 \\
\bottomrule
\end{tabular}
\end{table}
 
\begin{figure*}[!t]
	\begin{minipage}[t]{0.49\textwidth}
		\centering
		\includegraphics[width=0.88\linewidth]{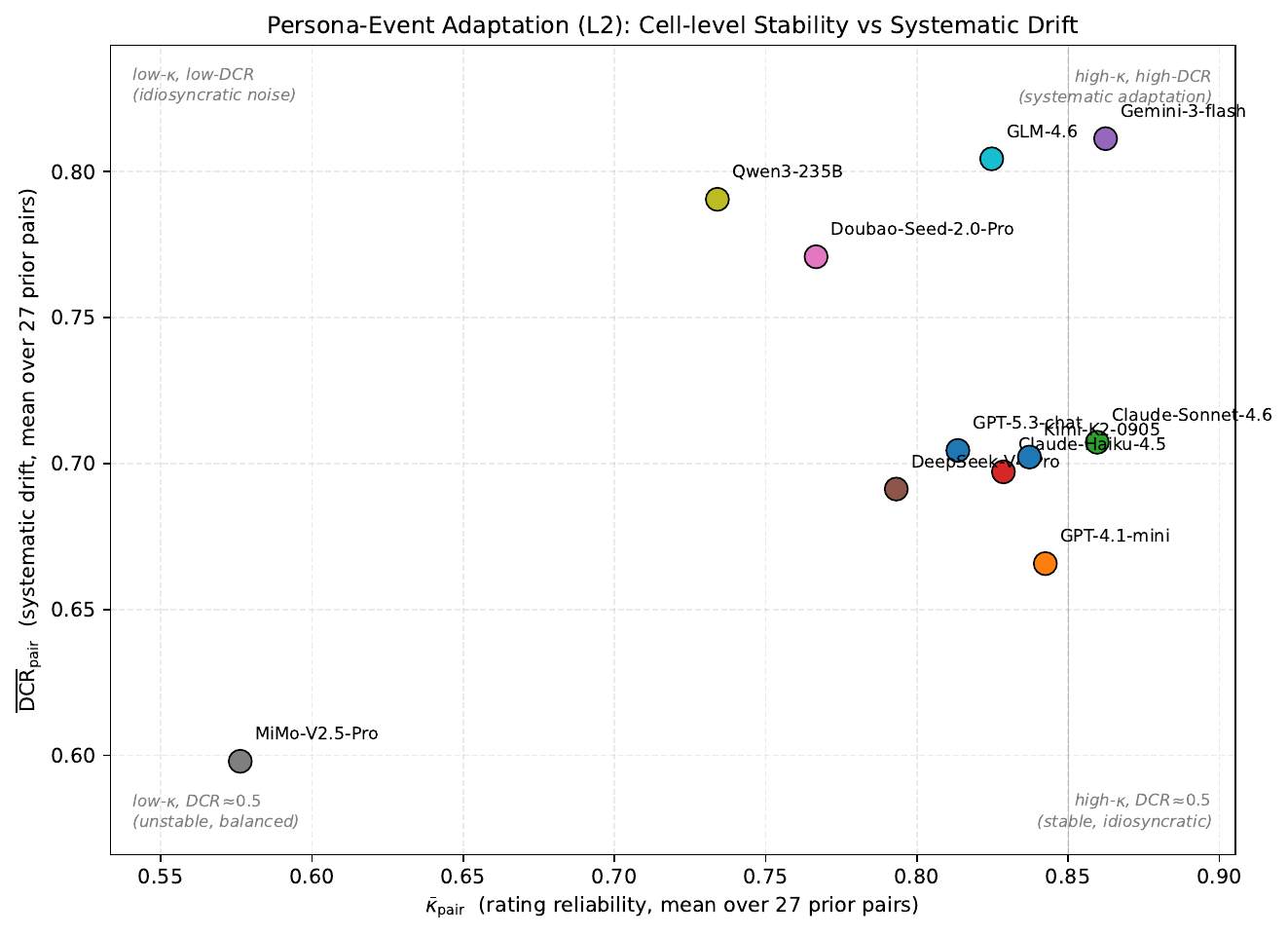}
		\caption{The $(\bar{\kappa}_{\text{pair}}, \overline{\mathrm{DCR}}_{\text{pair}})$ plane for the 11-model PC-Agent benchmark. Averages use the 27 definite-direction pairs.}
		\label{fig:kappa_dcr_plane}
	\end{minipage}\hfill
	\begin{minipage}[t]{0.49\textwidth}
		\centering
		\includegraphics[width=0.75\linewidth]{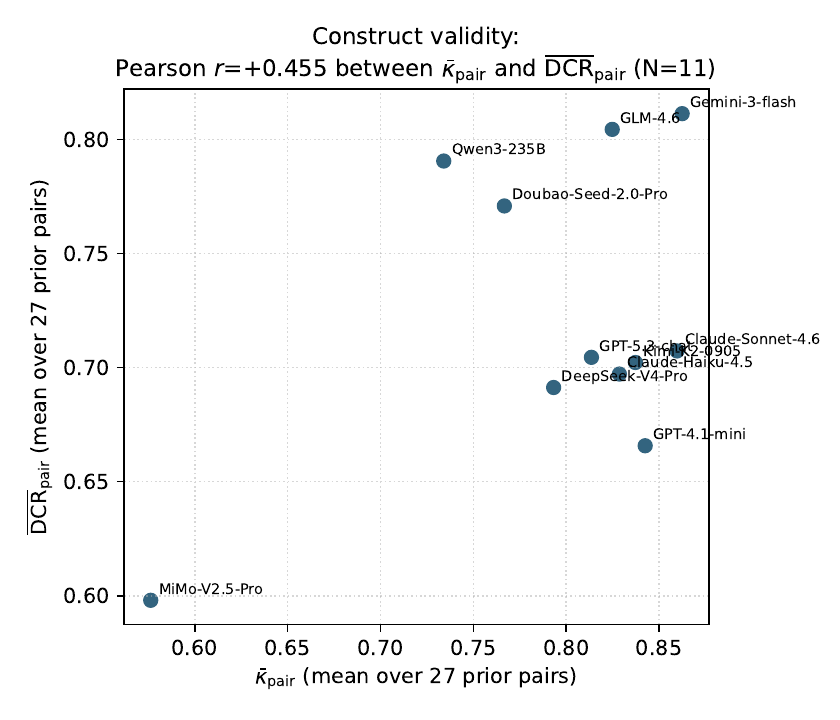}
		\caption{Construct validity of the diagnostic. Pair-level Pearson correlation is $r(\bar{\kappa}_{\text{pair}},\overline{\mathrm{DCR}}_{\text{pair}}){=}{+}0.455$ across 11 models. The two measures remain separable with a modest positive correlation.}
		\label{fig:construct_validity}
	\end{minipage}
\end{figure*}

\begin{figure*}[!t]
	\begin{minipage}[t]{0.49\textwidth}
		\centering
		\includegraphics[width=0.95\linewidth]{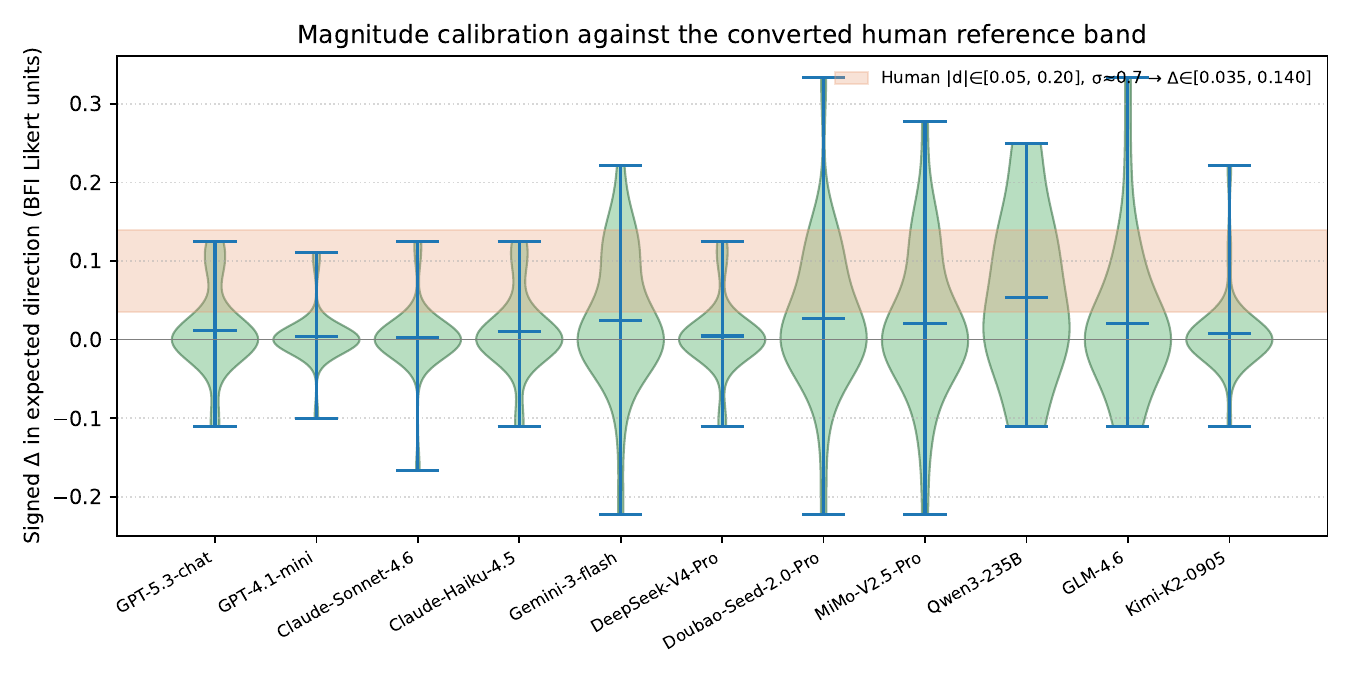}
		\caption{RQ2 persona-level median $\Delta$ for each definite-direction event--trait pair. The shaded band converts the representative standardized human effect-size band $|d|\in[0.05,0.20]$ to BFI Likert units using $\sigma{\approx}0.7$, yielding $\Delta\in[0.035,0.14]$ \citep{buhler2024life,bleidorn2018life}.}
		\label{fig:rq2_magnitude}
	\end{minipage}\hfill
	\begin{minipage}[t]{0.49\textwidth}
		\centering
		\includegraphics[width=0.98\linewidth]{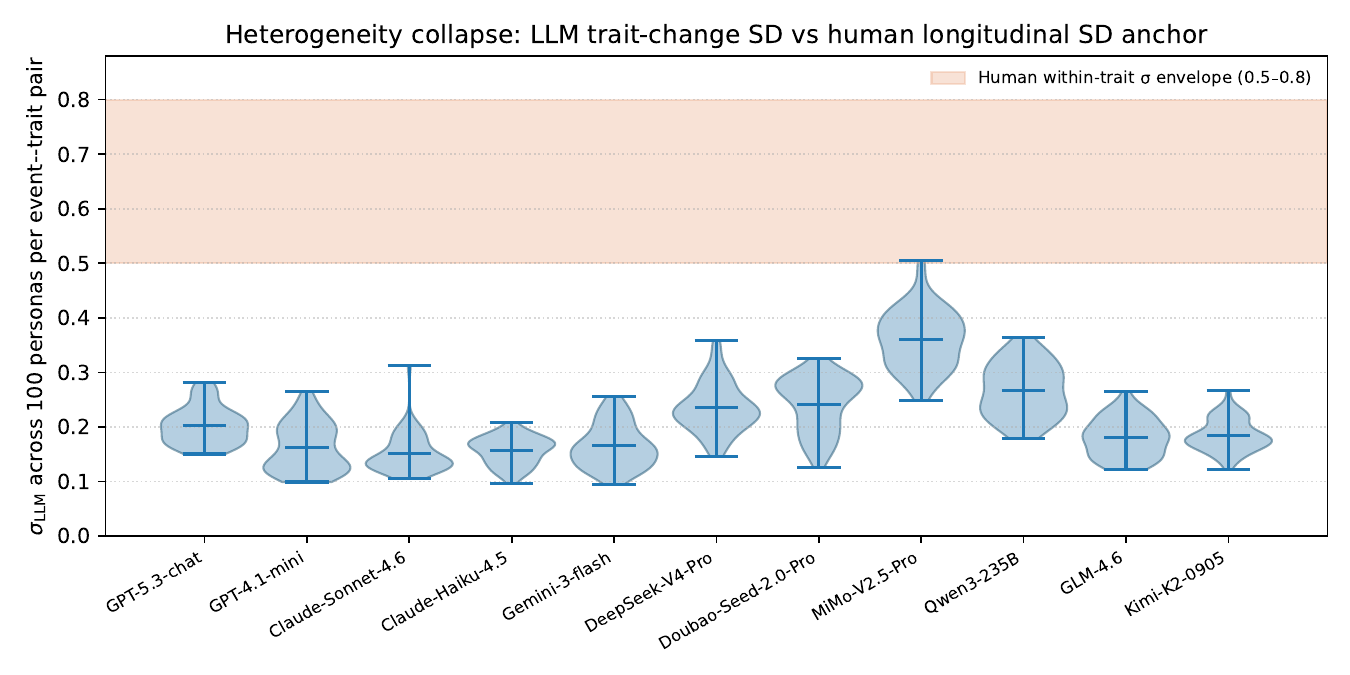}
		\caption{RQ4 distribution of $\sigma_\text{LLM}$ across 100 personas for each model and event--trait pair. The shaded band shows the human within-trait SD envelope \citep{buhler2024life,roberts2006patterns}.}
		\label{fig:rq4_sigma}
	\end{minipage}
\end{figure*}

\begin{figure*}[!t]
	\centering
	\includegraphics[width=0.68\textwidth]{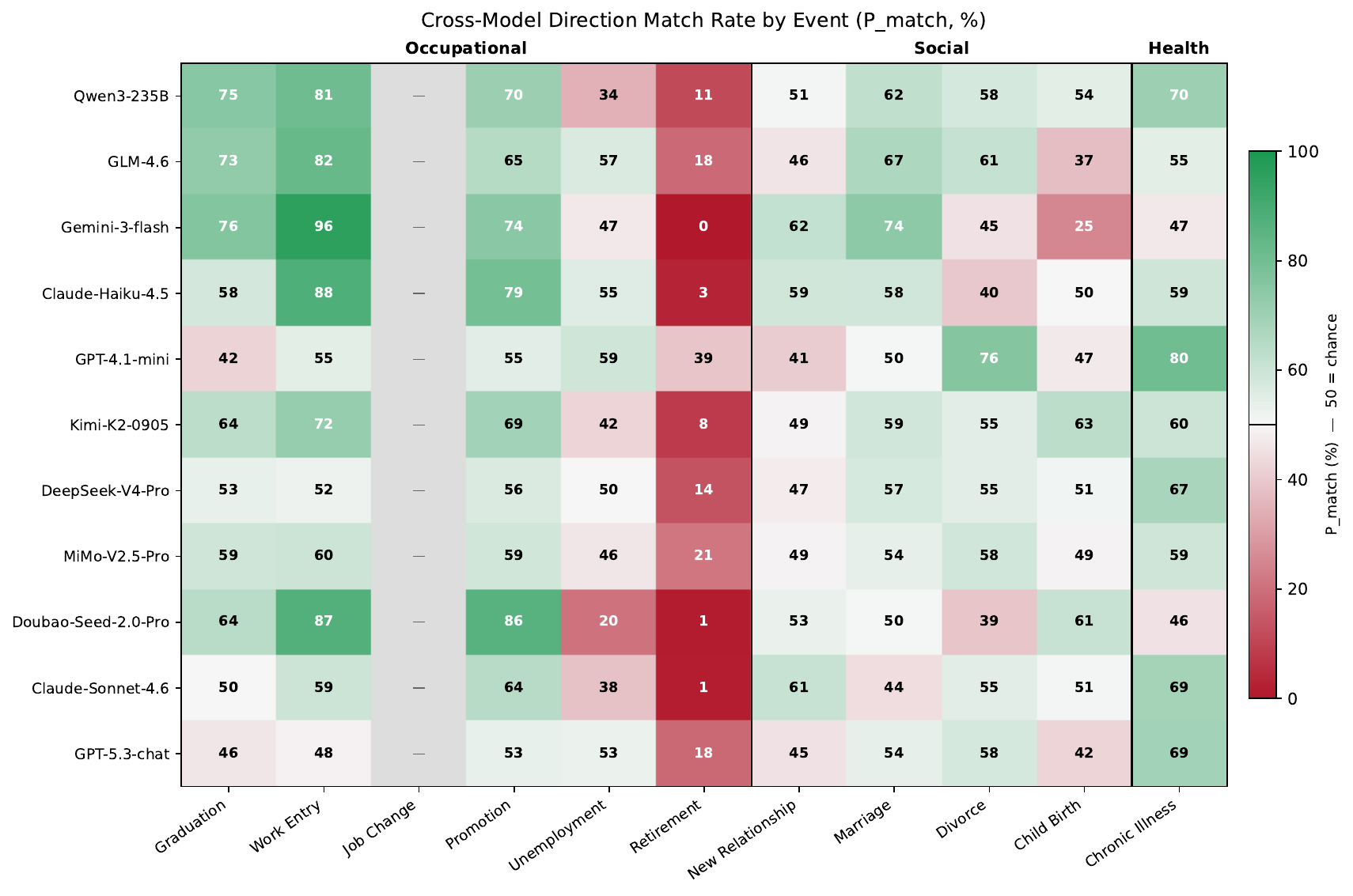}
	\caption{Direction match rate $P_\text{match}$ (\%) for each model and event combination. Red marks values below 50, and green marks values above 50. Occupational onboarding and chronic illness carry the aggregate signal. Retirement is universally inverted.}
	\label{fig:match_heatmap}
\end{figure*}

\end{document}